\pdfoutput=1

\documentclass[11pt]{article}

\usepackage[final]{acl}

\usepackage{times}
\usepackage{latexsym}
\usepackage{amsfonts}
\usepackage{amsmath}
\usepackage{amssymb}
\usepackage{mathrsfs}
\usepackage{calligra}
\usepackage{tikz}
\usepackage{multirow}
\usepackage{tcolorbox}
\tcbuselibrary{breakable}
\usepackage{booktabs}
\usepackage{xspace}
\usepackage{enumitem}
\usepackage{listings}
\usepackage{mathtools}
\usepackage[T1]{fontenc}

\usepackage[utf8]{inputenc}

\usepackage{microtype}

\usepackage{inconsolata}

\usepackage{graphicx}
\usepackage{verbatim}
\newcommand{\ours}{\textsc{ATLaS}\xspace}

\title{\ours: Agent Tuning via Learning Critical Steps}

\author{
 \textbf{Zhixun Chen\textsuperscript{1}\protect\footnotemark[1]},
 \textbf{Ming Li\textsuperscript{2}\protect\footnotemark[1]},
 \textbf{Yuxuan Huang\textsuperscript{3}},
 \textbf{Yali Du\textsuperscript{4}},
 \textbf{Meng Fang\textsuperscript{3}},
 \textbf{Tianyi Zhou\textsuperscript{2}}
\\
 \textsuperscript{1}University of Technology Sydney,
 \textsuperscript{2}University of Maryland,
 \\
 \textsuperscript{3}University of Liverpool,
 \textsuperscript{4}King's College London,
 \\
\texttt{zhixun.chen@student.uts.edu.au, \{minglii, tianyi\}@umd.edu} \\
\texttt{\{yuxuan.huang, meng.fang\}@liverpool.ac.uk, yali.du@kcl.ac.uk}
}

\begin{document}
\maketitle
\renewcommand{\thefootnote}{}
\renewcommand{\thefootnote}{\arabic{footnote}}
\begin{abstract}
Large Language Model (LLM) agents have demonstrated remarkable generalization capabilities across multi-domain tasks. Existing agent tuning approaches typically employ supervised finetuning on entire expert trajectories. 
However, behavior-cloning of full trajectories can introduce expert bias and weaken generalization to states not covered by the expert data. 
Additionally, critical steps—such as planning, complex reasoning for intermediate subtasks, and strategic decision-making—are essential to success in agent tasks, so learning these steps is the key to improving LLM agents. 
For more effective and efficient agent tuning, we propose \ours that identifies the critical steps in expert trajectories and finetunes LLMs solely on these steps with reduced costs. By steering the training's focus to a few critical steps, our method mitigates the risk of overfitting entire trajectories and promotes generalization across different environments and tasks. 
In extensive experiments, an LLM finetuned on only $\sim$30\% critical steps selected by \ours outperforms the LLM finetuned on all steps and recent open-source LLM agents. \ours maintains and improves base LLM skills as generalist agents interacting with diverse environments. \looseness-1
\end{abstract}

\section{Introduction}

\begin{figure}[ht]
    \centering
    \includegraphics[width=0.9\linewidth]{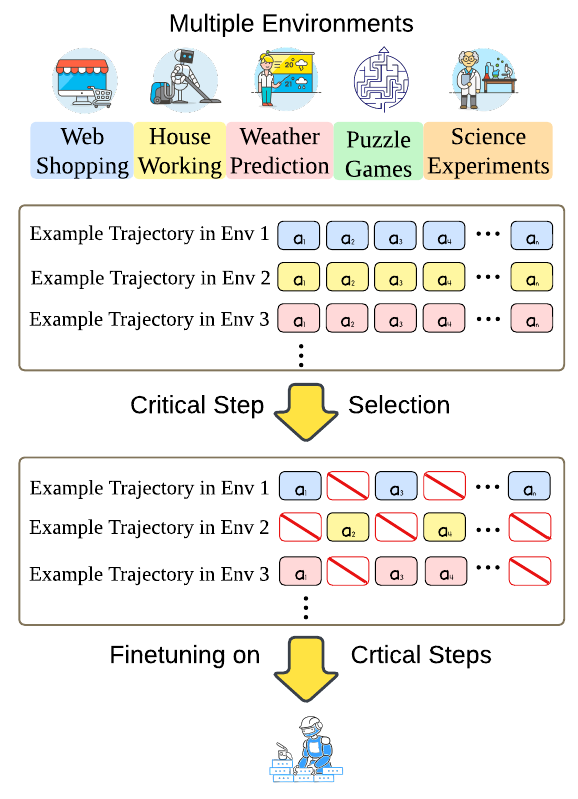}
    \caption{The proposed \ours identifies the critical steps in expert trajectories collected from diverse interactive environments and finetunes the agent on these steps only, where $a_i$ represents the expert action in step-$i$. \ours alleviates the potential overfitting to experts' every-step behaviors and achieves better generalizability by training on much fewer steps (``less is better'').} 
    \vspace{-1em}
    \label{fig:intro}
\end{figure}
An intelligent agent can perceive its environment, process information, and take actions to achieve specific goals or objectives \citep{maes1995agents, wooldridge1995intelligent}. Traditional AI agents, especially empowered by Monte Carlo Tree Search (MCTS) or Reinforcement learning, have shown great potential in complex and difficult tasks \citep{silver2017mastering, vinyals2019grandmaster}. However, these methods are task-specific and suffer from weak generalization across different task domains or environments. 
On the other hand, the recent rise of Large Language Models (LLMs) \cite{ NEURIPS2020_1457c0d6, openai2024gpt4technicalreport, touvron2023llama, touvron2023llama2, jiang2023mistral} provide powerful foundation models with rich prior knowledge and reasoning capabilities, leading to remarkable generalization performance across 
various downstream applications \cite{zhao2023survey, Xu2024ASO}. 
Their unprecedented performance in different domains, including instruction following, reasoning, planning, and tool usage, makes LLMs ideal agents in multi-task settings \cite{wei2022chain, kojima2022large, qin2023toolllm, patil2023gorilla, li2023reflectiontuning, li-etal-2024-selective, li2024mosaic, wang2024survey}. 
\looseness-1

Despite their effectiveness, transferring capabilities across tasks remains challenging due to overfitting to expert trajectories and training inefficiency. Most existing agent tuning methods rely on imitation of every step of expert trajectories \cite{chen2023fireact, zeng2023agenttuning, chen2024agent, zhang2024agentohana, xi2024agentgym, song2024agentbank}. For instance, FireAct \cite{chen2023fireact} generates diverse expert trajectories to fine-tune LLMs with enhanced planning abilities, while AgentTuning \cite{zeng2023agenttuning} combines expert trajectories from various interactive environments with general instruction datasets to improve agents' generalization capability.\looseness-1

However, fine-tuning on the entire expert trajectories can introduce expert bias, causing the model to overfit to specific behaviors and diminishing its ability to generalize to unseen environments \cite{ghosh2024closer}. While experts in a few critical steps indeed provide key supervised information to improve the agent, the agent may already excel in other non-critical steps. Additionally, fitting the full trajectories of one task may lead to declines in others due to the potential distribution gap and negative transfer between tasks \cite{song2024agentbank}. Moreover, imitating expert trajectories on redundant steps or sub-optimal/replaceable actions incurs excessive training costs. 
\looseness-1

\looseness-1



To address the above challenges, we propose \textbf{\ours}, \textbf{\underline{A}}gent \textbf{\underline{T}}uning via \textbf{\underline{L}}earning Critic\textbf{\underline{a}}l \textbf{\underline{S}}teps, 
a novel approach that identifies the critical steps within expert trajectories and only finetunes LLMs on those selected steps, which is illustrated in Figure~\ref{fig:intro}. 
By steering the training's focus to a few critical steps, we not only reduce the backpropagation cost for training but also lower the overfitting risk to whole trajectories. Moreover, imitation on partial trajectories mitigates the expert bias, and encourages generalization across environments and tasks.\looseness-1

In \ours, an oracle LLM (GPT4o by default) is utilized as the selector to identify the critical steps semantically based on four criteria: key observations, plan formulation, recalling prior information, and pivotal actions. These critical steps are essential to the success of downstream tasks, but they are usually challenging to be generated by the base LLMs. 
Our training focuses on finetuning base models on these selected steps. 
\looseness-1
\begin{figure}
    \centering
    \includegraphics[width=1.0\linewidth]{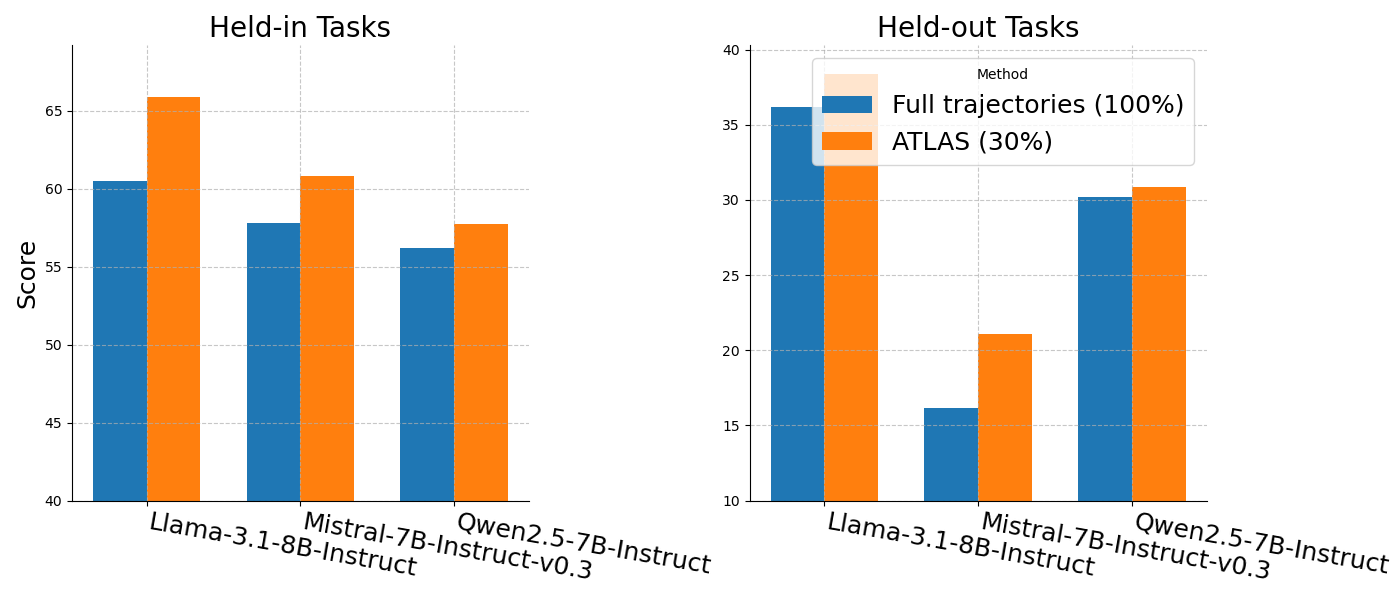}
    \caption{Three base LLMs finetuned by \ours vs. full trajectories (100\% of the steps), evaluated on held-in and held-out agentic tasks. \ours consistently outperforms full-trajectory finetuning, indicating better generalizability of \ours by training on fewer but critical steps.\looseness-1} 
    \label{fig:bar_dif_model}
\end{figure}
Our experiments provide extensive comparisons between \ours and other agent-tuning strategies or LLM agents. As indicated by the results in Figure~\ref{fig:bar_dif_model},  
\ours-finetuned agent maintains and improves the generalization capabilities of base LLMs in diverse environments, excelling on both held-in and held-out tasks. 
Our main contributions can be summarized as:  

\begin{itemize}  
    \item We introduce a novel method \ours reducing the tokens for agent tuning to $30\%$, by selecting the critical steps in expert trajectories. \looseness-1
    \item Agents finetuned on $30\%$ critical steps outperforms the baseline agents finetuned on $100\%$ steps, especially in the multi-task learning scenario, mitigating the expert bias and negative transfer across tasks. 
    \item \ours-finetuned agent achieves a better generalization capability than baseline agents on not only held-in but also held-out tasks. \looseness-1
\end{itemize}  


\begin{figure*}
    \centering
    \includegraphics[width=1.0\linewidth]{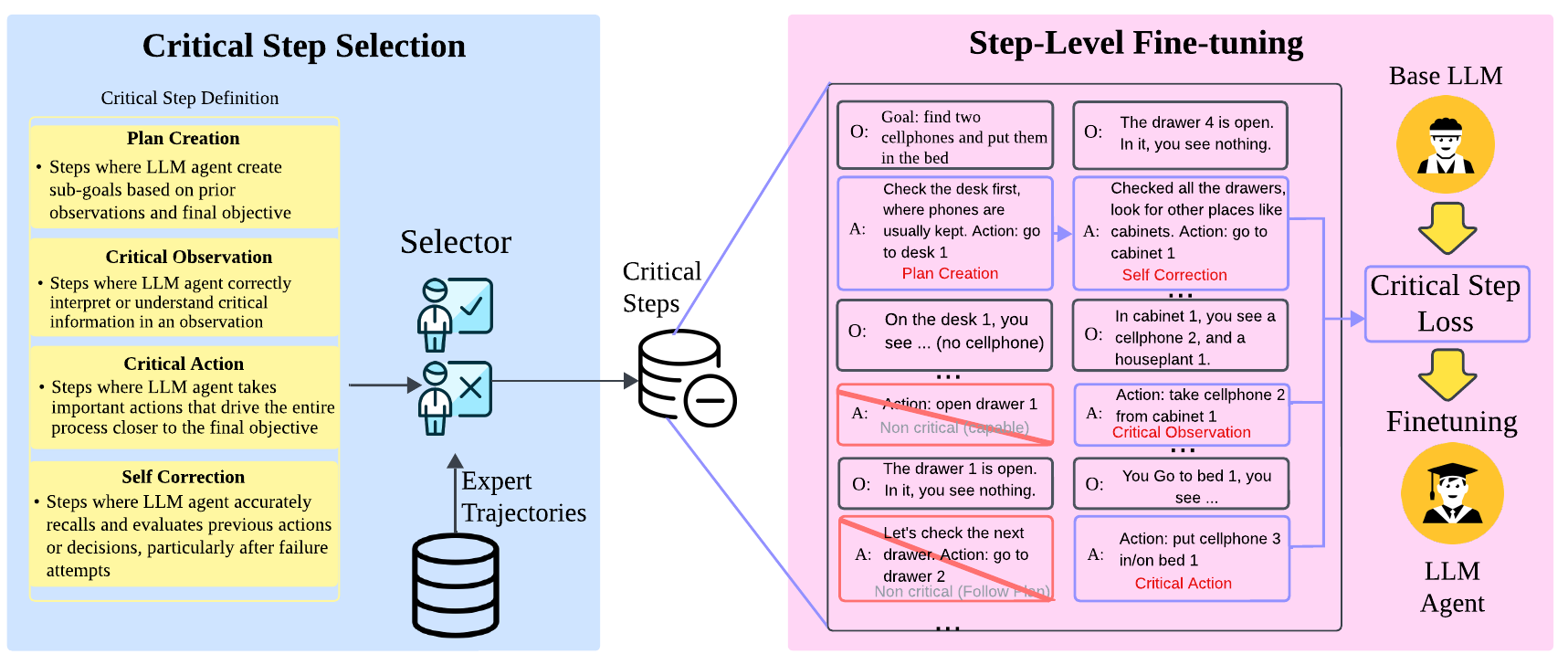}
    \vspace{-1.5em}
    \caption{Overall of \ours. The selector identifies critical steps in expert trajectories collected in multiple environments, where ``O'' and ``A'' denote observation and action, respectively. Training loss is only computed on the critical steps. This encourages more exploration of non-critical steps, reduces the training cost, and improves the agent's generalization performance.} 
    \vspace{-1.em}
    \label{fig:illustration}
\end{figure*}
 


\section{Preliminaries}
As a multi-task learning framework, we define a set of $N$ environments as $\mathcal E$. For any particular environment $E \in \mathcal E$ and a task prompt space $I$, the agent's dynamics is formalized as a Partially Observable Markov Decision Process (POMDP) \citep{sutton2018reinforcement} $\mathcal M_{E,I}\triangleq(S, A, O, T, r)$. In this formalization, $S$ denotes the state space, $A$ denotes the action space, $O$ denotes the observation space, $T: S \times A \to S$ represents the state transition function, and $r: S \times A \to \mathbb{R}$ defines the reward function.
\looseness-1

Upon receiving an instruction $i \in I$ in environment $E$, the agent takes action $a_t \sim \pi_\theta(\cdot |o_t;E,i)$ according to its policy $\pi_\theta$, where $\theta$ represents the policy parameters. For LLM agents, $\pi_\theta$ is an LLM, and its output is not limited to actions and depends on the prompt. The environment then transitions to a new state $s_{t+1} \in S$ based on the action, and the agent receives a new observation $o_{t+1} \in O$. This interaction continues as the agent engages with the environment until the task is completed or the maximum step limit is reached. We adopt the ReAct framework \citep{yaoreact} for the LLM agent: at each step $t$, the agent first generates reasoning thought $h_t$ with a think prompt $\texttt{prompt}_h$. Define the trajectory up to step-$t$ as
\begin{equation}
\tau_{1:t} \triangleq (h_1, a_1, o_1, \dots, h_t, a_t, o_t),
\end{equation}
The thought $h_t$ is sampled by
\begin{equation}
    h_{t} \sim \pi_\theta(\cdot |\tau_{1:t-1};\texttt{prompt}_h,E,i).
\end{equation}
The action is then drawn from the same model with another prompt $\texttt{prompt}_a$, i.e.,
\begin{equation}
    a_{t} \sim \pi_\theta(\cdot |\tau_{1:t-1},h_t;\texttt{prompt}_a,E,i).
\end{equation}
The return of trajectory $\tau=\tau_{1:T}$, $ G(\tau) = \sum_{t=1}^{T} \gamma^t r_t $ with a discounting factor $ \gamma $, reflects the agent's performance in task $ i $ within environment $ E $. We collect a dataset $D$ of expert trajectories for all environments in $\mathcal E$ and tasks in $I$.

\section{Methodology}
In this section, we introduce our \ours, which identifies critical steps of the expert trajectories and finetunes LLM agents solely on them. 
\looseness-1

\subsection{Definition of Critical Steps}
Critical steps are key actions or states within an agent’s trajectory that have a substantial impact on the return $ G(\tau)$. As depicted in Figure \ref{fig:illustration}, these steps are essential for ensuring the success of the trajectory. We characterize critical steps as those requiring precise decision-making and flawless execution by the agent. In contrast, non-critical steps are more flexible, allowing for adjustments or rearrangements without compromising the overall trajectory. Typically, critical steps are challenging for the base model's policy to generate, whereas non-critical steps are relatively easy for the base model. To identify these steps, we sample expert trajectories with dense rewards in available environments. We then empirically categorize four types of critical steps: \textbf{Plan Creation}, \textbf{Critical Observation}, \textbf{Critical Action}, and \textbf{Self Correction}.
\subsection{Selection of Critical Steps}
Properly identifying and extracting critical steps from a trajectory is essential for ensuring effective analysis and learning. However, unlike instance-level data selection for LLMs, which has been explored \cite{li-etal-2024-quantity, li-etal-2024-superfiltering}, step-level selection is challenging since the task-related critical steps can differ dramatically by their semantic meaning.
To address this issue, we leverage the selector to identify critical steps based on the critical step identification prompt, $\texttt{prompt}_c$. This approach ensures a more nuanced and context-aware assessment, helping to accurately identify the most influential steps for optimal performance. 

The prompt initially defines the critical steps with the four categories introduced above:
\begin{itemize}
\setlength{\itemsep}{2pt}
\setlength{\parsep}{2pt}
\setlength{\parskip}{2pt}
    \item \textbf{Plan Creation}: Steps where the LLM agent formulates sub-goals by analyzing previous observations and considering the final objective, breaking down the larger goal into manageable tasks that guide the agent's actions toward the overall outcome.
    \item \textbf{Critical Observation}: Steps where the LLM agent identifies and analyzes key information from the environment, which helps the agent understand the objective or state and refine its strategy and decision-making toward more effective outcomes.
    \item \textbf{Critical Action}: Steps where the LLM agent takes decisive and impactful actions based on prior observations, significantly advancing the process toward the final objectives. These actions are crucial in shaping the direction of the agent's strategy and are often pivotal moments that determine progress or failure, ensuring that the agent remains on track to achieve the desired outcome.
    \item \textbf{Self Correction}: Steps where the LLM agent carefully recalls and assesses its previous actions or decisions, especially after encountering failure or suboptimal outcomes. During this process, the agent reflects on what went wrong, identifies areas for improvement, and adjusts its approach to enhance future performance, which helps the agent refine its decision-making and better align with the overall objective.
    \looseness-1
\end{itemize}
\vspace{-1cm}
After defining critical steps, $\texttt{prompt}_c$ guides the selector to comprehend the given trajectory and summarize a high-level plan with sub-goals of the trajectory to enhance its understanding of the trajectory. Based on the sub-goals, the selector identifies the critical steps and categorizes them into four predefined categories. Notably, we require the selector to select at most \( m \) percentage of steps in the trajectory and return their indices as\looseness-1 
\begin{equation}
    C_\tau = \texttt{selector}(\tau; \texttt{prompt}_c, m)
\end{equation}
\ours does not ignore all other steps in the trajectory during training. Instead, we keep the whole original trajectory, including the non-critical steps, as the input sequence but do not compute teacher-force loss on their tokens during training. \looseness-1
\begin{tcolorbox}[title = {Critical Step Identification Prompts},boxsep=1pt, left=2pt, right=2pt, top=2pt, bottom=2pt, fonttitle=\small, fontupper=\small]
\begin{enumerate}[leftmargin=4mm]
    \item Generate a high-level plan or strategy from the expert conversation, summarizing the key sub-goals required to complete the task.
    \item Determine the most critical action steps in the expert conversation based on the plan and sub-goals, ensuring the number of selected steps does not exceed \{$m\% \times length(\tau)$\}.
    \item Justify your selection of these critical steps, specifying the category to which each step belongs.
\end{enumerate}
 \end{tcolorbox}
The full prompts are provided in Appendix \ref{a_c_prompt}.

\subsection{Agent Tuning on Critical Steps}
Using the critical step dataset \( D_c \), we optimize the model by calculating the loss exclusively on these critical steps. Training the agent on this reduced set of essential steps enhances both the efficiency and effectiveness of the fine-tuning process, ensuring the model focuses on the key actions that contribute to task success and gains knowledge rather than merely mimicking expert behavior. Thus, the objective function can be defined as:
\begin{equation}
\notag
\begin{split}
\hspace{-0.5em}
  \mathcal{J}(\theta) = 
  \displaystyle \mathop{\mathbb{E}}_{\tau\sim D} \Biggl[ &\sum_{t\in C_\tau} \log \pi_\theta (h_t | \tau_{1:t- 1};\texttt{prompt}_h, E,i ) + \\
  & \log \pi_\theta (a_t | \tau_{1:t- 1},h_t;\texttt{prompt}_a, E,i ) \Biggr],
\end{split}
\end{equation}
where $\theta$ denotes the trainable parameter of the LLM agent.
This expectation-based objective ensures that the model maximizes probabilities at the critical steps that most influence task success, following the same reasoning as in the full-trajectory approach but applied selectively to $D_c$.\looseness-1

\section{Experiments} \label{sec: experiment}
\subsection{Environments}
In our experimental framework, environments are categorized into \textbf{held-in} and \textbf{held-out} sets to evaluate the model's performance. \textbf{Held-in environments} refer to the set of environments that are included in the training dataset. These are the environments in which the models can be fine-tuned, allowing them to learn and adapt their behavior based on the provided expert trajectories. The performance of the held-in environments measures how well the model has learned from the training data and its ability to replicate or improve upon expert behaviors within familiar settings. On the other hand, \textbf{held-out environments} consist of environments and tasks that are excluded from the training process. These environments are used to assess the model's generalization capabilities, determining how effectively it can apply learned knowledge to novel and unseen scenarios. The performance of held-out environments indicates the model's ability to generalize its decision-making and problem-solving skills beyond the specific examples it was trained on, highlighting its potential for adaptability in diverse and dynamic real-world applications.

Specifically, our held-in environments include web navigation task Webshop \citep{yao2022webshop}; textual digital games Textcraft \citep{prasad2023adapt}, Wordle and Maze \citep{abdulhai2023lmrl}; embodied games Alfworld \citep{shridhar2020alfworld}, Scienceworld \citep{wang2022scienceworld} and Babyai \citep{chevalier2018babyai}; tool using tasks Weather, Todo and Movie \citep{ma2024agentboard}. It is important to note that some environments provide a dense reward $r \in [0,1]$, while others offer only sparse rewards $r \in \left\{0,1\right\}$. To maintain simplicity and consistency, we follow the evaluation matrix definition used in previous work \citep{singh2024humandatascalingselftraining,xi2024agentgym} across all tasks. For the held-out environments, we apply two tool usage environments Sheet and Academic, and two game environments Jericho and PDDL \cite{ma2024agentboard}. A detailed description of these environments can be found in Appendix \ref{env_detail}.

\subsection{Implementation Details}
We apply the open-sourced dataset AgentTraj-L \footnote{https://huggingface.co/datasets/AgentGym/AgentTraj-L} \citep{xi2024agentgym} as our unfiltered dataset $D_o$, which contains the expert trajectories of all held-in tasks mentioned above. We then filter the dataset to obtain the critical step dataset $D_c$ by prompting GPT-4o with the selection instruction $I_c$. For more details, please refer to Appendix \ref{a_c_prompt}. 

We employ Llama-3.1-8B-Instruct \citep{dubey2024llama} as the backbone model for our agent unless specified. During training, we focus on the critical step $D_c$, computing loss exclusively at these pivotal steps in the trajectory. We set the maximum selection ratio $m$ for the critical step to be $30\%$. More information about the training setup is provided in Appendix \ref{train_conf}. The evaluation prompts are based on the React format \citep{yaoreact}, consistent with the structure of the training data.
\begin{table*}[t]
\centering
\resizebox{\textwidth}{!}{ 
\begin{tabular}{l|ccccccccccc|ccccc}
\hline
\multirow{2}{*}{\textbf{Model}} & \multicolumn{11}{|c|}{\textbf{held-in}} & \multicolumn{5}{|c}{\textbf{held-out}} \\
\cline{2-17}
& Alfworld & Babyai & Maze & Movie & Sciworld & Textcraft & Todo & Weather & Webshop & Wordle & AVG & Academic & Sheet & Jericho & PDDL & AVG \\ 
\hline
\multicolumn{14}{l}{\textit{Closed-Source Model}} \\ 
\hline
DeepSeek-Chat & 51.00 & 45.67 & 4.00 & 70.00 & \textbf{16.80} & 23.00 & 75.00 & 70.00 & 11.00 & 24.00 & 39.05 & - & - & - & - & - \\
Claude-3-Haiku & 0.00 & 1.93 & 4.00 & 50.00 & 0.83 & 0.00 & 65.00 & 55.00 & 5.50 & 16.00 & 19.83 & - & - & - & - & -\\
Claude-3-Sonnet & 13.00 & 79.25 & 4.00 & 50.00 & 2.78 & 38.00 & 80.00 & 65.00 & 1.50 & 36.00 & 36.95 & - & - & - & - & - \\
GPT-3.5-Turbo & 26.00 & 71.36 & 4.00 & 70.00 & 7.64 & 47.00 & 40.00 & 25.00 & 12.50 & 20.00 & 32.35 & 65.00 & 38.34 & 19.93 & 24.98 & 37.06\\
GPT-4-Turbo & \textbf{67.50} & \textbf{72.83} & \textbf{68.00} & \textbf{95.00} & 14.38 & \textbf{77.00} & \textbf{95.00} & \textbf{80.00} & \textbf{15.50} & \textbf{88.00} & \textbf{67.32} & \textbf{80.00} & \textbf{75.83} & \textbf{52.44} & \textbf{81.16} & \textbf{72.36}\\
\hline
\multicolumn{14}{l}{\textit{Open-Source Base Model}} \\
\hline
Llama-2-7B-Instruct &0.00 & 23.00 & 0.00 & 0.00 & 7.20 & 0.00 & 0.00 & 0.00 & 0.00 & 0.00 & 3.02& 0.00 & 0.00 & 1.25 & 0.83 & 0.52\\
Llama-3-8B-Instruct &0.00 & 61.60 & 12.00 & 60.00 & 70.82 & \textbf{15.00} & 60.00 & 30.00 & \textbf{13.00} & 0.00 & \textbf{32.24} & 40.00 & 35.40 & 10.51 & \textbf{16.67} & 25.65\\
Llama-3.1-8B-Instruct &0.00 & 60.69 & \textbf{16.00} & 65.00 & 19.57 & 9.00 & 65.00 & \textbf{40.00} & 2.00 & 0.00 & 27.72& 40.00 & 37.31 & \textbf{19.23} & 12.18 & \textbf{27.18} \\
Phi-3-mini-128k-instruct &0.00 & 19.80 & 8.00 & 50.00 & 19.10 & 1.00 & 40.00 & 20.00 & 1.00 & 4.00 & 16.29& 25.00 & 30.21 & 1.25 & 1.25 & 14.43\\
Phi-3.5-mini-instruct &0.00 & 21.00 & 12.00 & 60.00 & 18.20 & 3.00 & 60.00 & 30.00 & 2.00 & 6.00 & 21.22& 30.00 & 33.62 & 0.56 & 0.83 & 16.25\\
Gemma2-9B-it &0.00 & 52.00 & 8.00 & 55.00 & \textbf{72.60} & 0.00 & 45.00 & 0.00 & 0.00 & 0.00 & 23.26 & 30.00 & 34.46 & 14.02 & 7.20 & 21.42\\
Mistral-7B-Instruct-v0.3 &0.00 & 17.30 & 4.00 & 0.00 & 48.00 & 0.00 & 10.00 & 5.00 & 0.00 & 0.00 & 8.43 & 0.00 & 9.69 & 1.96 & 2.50 & 3.54\\
Qwen2.5-7B-Instruct & 0.00 & \textbf{67.00} & \textbf{16.00} & \textbf{75.00}  & 15.30 & 7.00 & \textbf{85.00} & 30.00 & 3.00 & \textbf{12.00} & 31.03 & \textbf{55.00} & \textbf{39.00} & 14.26 & 0.28 & 27.14 \\
\hline
\multicolumn{14}{l}{\textit{Fine-tuned Agents}} \\
\hline
Xlam-7B-r & 17.50 & 62.00 & 0.00 & 20.00  & 12.00& 1.00 & 15.00 & 10.00 & 37.50 & 4.00 & 17.9 & 35.00 & 28.40 & 8.92 & 3.06 & 18.85\\
AgentLM-7B & 71.00 & 0.49 & 12.00 & 5.00 & 1.63 & 4.00 & 15.00 & 0.00 & 36.50 & 4.00 & 14.96 & 10.00 & 13.30 & 15.88 & 2.78 & 10.49\\
AgentEvol-7B & \textbf{88.00} & \textbf{82.70} & 12.00 & 60.00 & 38.00 & 64.00 & 70.00 & 25.00 & \textbf{76.50} & 12.00 & 52.82 & 25.00 & 26.20 & 4.05 & 3.10 & 14.59\\
AgentTraj-L (100\% steps) & 83.00&	70.31&	\textbf{48.00}&	75.00 &	37.92&	71.00&	85.00&	55.00&	68.00&	12.00&	60.52 & \textbf{75.00} & 42.83 & 15.18 & 11.71 & 36.18 \\
Perplexity Selection & 78.50 & 77.59 & 24.00 & 65.00 & \textbf{53.54} & 59.00 &70.00 & 30.00 & 66.50 &8.00 & 53.21 & 30.00 & 36.48 & 13.28 & 8.76 & 22.13 \\
\ours (30\% steps) & 84.50&	80.98&	\textbf{48.00}&	\textbf{90.00}&	42.02&	\textbf{72.00}&	\textbf{90.00}&	\textbf{60.00}&	71.50&	\textbf{20.00}&	\textbf{65.91} & 70.00 & \textbf{49.39} & \textbf{18.21} & \textbf{15.84} & \textbf{38.36}\\
\hline
\end{tabular}
}
\caption{Performance of \ours-finetuned LLM agents vs. baseline LLMs and finetuned agents on held-in and held-out test tasks. ``Unfilter'' finetunes an LLM on the unfiltered dataset of complete expert trajectories. The best performance in each category is highlighted in \textbf{bold}. By finetuning LLMs only on the critical steps of expert trajectories, \ours achieves outstanding agent performance on both held-in and held-out tasks. \looseness-1}
\label{tab: main results}
\end{table*}
\subsection{Baselines}
We compare our LLM agent against three types of baselines: closed-source models, open-source models, and other fine-tuned LLM agents. Specifically, for closed-source models, we include GPT-3.5-Turbo \citep{ouyang2022training}, GPT-4-Turbo \citep{openai2024gpt4technicalreport}, Claude 3 \citep{antro2024claude}, and DeepSpeek-Chat \citep{liu2024deepseek}. For open-source models, we evaluate Llama3-8B-Instruct, Llama3.1-8B-Instruct \citep{dubey2024llama}, Gemma-2-9b-it \citep{gemma_2024}, Phi-3-mini-128k-instruct, Phi-3.5-mini-instruct \citep{phi3}, Mistral-7B-Instruct-v0.3 \cite{jiang2023mistral} and Qwen2.5-7B-Instruct \citep{qwen2}. For LLM agents fine-tuned on expert trajectories, we compare Xlam-7B-r \citep{zhang2024agentohana}, AgentLM from AgentTuning\citep{zeng2023agenttuning} and AgentEvol from AgentGym \citep{xi2024agentgym}. We also fine-tune AgentTraj-L using the same base model, but with SFT on the unfiltered dataset \( D_o \) as a baseline for a fair comparison. Additionally, we introduce another baseline where the base model's perplexity on all steps of the training dataset is used to select critical steps. Specifically, we choose the top 30\% of steps with the highest perplexity and fine-tune on those steps.

\subsection{Main Results}

The results shown in Table \ref{tab: main results} demonstrate the performance of various models on held-in and held-out tasks. For closed-source models, the reported outcomes follow those presented in AgentGym \cite{xi2024agentgym}. On held-in tasks, our approach, fine-tuned on just 30\% of the entire tokens of the expert trajectories, outperforms baseline methods across most tasks. It also achieves over a 5\% average improvement compared to models fine-tuned with unfiltered data. Furthermore, in certain tasks such as Alfowrld and Babyai, our model surpasses some advanced closed-source models. The perplexity-based selection performs poorly because it solely measures the token generation difficulty of one step by the base model, which is affected by both the step's importance and expert bias. Consequently, using perplexity for selection forces the model to mimic the expert's distribution rather than effectively learning the task knowledge to solve the problem.

The results on held-out tasks demonstrate that fine-tuning exclusively on critical steps enhances the generalization of agents to unseen tasks. Focusing solely on these steps reduces the need to fully replicate the expert LLM's distribution, allowing agents to learn the knowledge of how to solve problems effectively. Unlike other approaches, such as those described in \citep{zeng2023agenttuning}, which mix a large amount of general instruction data with task-specific trajectory data to enhance generalization, our method achieves comparable performance without relying on such datasets. This underscores the effectiveness of critical step training in boosting agents' performance on held-out tasks.

To validate the effectiveness of our approach across various backbone models, we perform experiments using Mistral-7B-Instruct-v0.3 \citep{jiang2023mistral} and Qwen2.5-7B-Instruct \citep{qwen2}. The results, summarized in Table \ref{fig:bar_dif_model}, demonstrate that our method consistently enhances performance across different models, yielding improvements on both held-in and held-out tasks. The detailed results are shown in Appendix \ref{app: add results}. 

\begin{table*}[t]
\centering
\resizebox{1.0\textwidth}{!}{ 
\begin{tabular}{l|ccccccccccc|ccccc}
\hline
\multirow{2}{*}{\textbf{Data}} & \multicolumn{11}{|c|}{\textbf{held-in}} & \multicolumn{5}{|c}{\textbf{held-out}} \\ 
\cline{2-17}
& Alfworld & Babyai & Maze & Movie & Sciworld & Textcraft & Todo & Weather & Webshop & Wordle & AVG & Academic & Sheet & Jericho & PDDL & AVG \\ 
\hline
Non-critical Steps (30\%) &78.00 & 70.70 & 32.00 & 75.00 & 39.02 &57.00 & 75.00 & 55.00 & 68.00 & 12.00 & 56.17 & 60.00 & 40.42 & 10.80 & 8.28 & 29.88  \\
Complete Trajectories (100\%) by AgentTraj-L & 83.00&	70.31&	\textbf{48.00}&	75.00&	37.92&	71.00&	85.00&	55.00&	68.00&	12.00&	60.52 & \textbf{75.00} & 42.83 & 15.18 & 11.71 & 36.18 \\
Critical Steps (30\%) by \ours & \textbf{84.50}&	\textbf{80.98}&	\textbf{48.00}&	\textbf{90.00}&	\textbf{42.01}&	\textbf{72.00}&	\textbf{90.00}&	\textbf{60.00}&	\textbf{71.50}&	\textbf{20.00}&	\textbf{65.91} & 70.00 & \textbf{49.39} & \textbf{18.21} & \textbf{15.84} & \textbf{38.36}\\
\hline
\end{tabular}
}
\caption{Critical vs. Non-critical steps \& complete trajectories for agent tuning (tested on held-in and held-out tasks). The agent trained on \ours-selected critical steps does not only outperform the agent trained on full trajectories but also surpasses the agent trained on non-critical steps by a large margin. This indicates that the critical steps improve agent tuning while training on non-critical steps might be detrimental to the generalization.}
\label{tab: non-critic}
\end{table*}
\subsection{Ablation Study}
\subsubsection{Critical Steps vs. Non-Critical Steps}
To assess the effectiveness of critical steps, we compare \ours with finetuning on non-critical steps \textbf{not} selected by the selector LLM in \ours. The number of steps for the non-critical selection dataset remains identical to our fine-tuning dataset, which is $30\%$ of the entire training dataset. 
\looseness-1

The experimental results, presented in Table \ref{tab: non-critic}, indicate that fine-tuning with critical steps positively impacts the performance on both held-in and held-out tasks. In contrast, fine-tuning with non-critical steps results in a negative impact across all tasks. These findings suggest that certain steps contribute positively to the agent's learning by providing useful knowledge, whereas other steps introduce negative biases, leading to a decline in performance.
\looseness-1

\subsubsection{Different Selection Ratio}
The ratio of critical steps plays a crucial role in the fine-tuning process. Therefore, we explore this hyperparameter and require the teacher model to select different proportions of steps from the expert trajectory as critical steps to identify the optimal ratio. As shown in Table \ref{tab:abl_random ratio}, a 30\% ratio yields the best agent performance on both held-in and held-out tasks. We also explore values of $m$ greater than 30\% but observed only a marginal increase in the number of critical steps selected by the selector. Consequently, we did not evaluate fine-tuning results for $m$ larger than 30\%. Motivated on these findings, we adopt $m = 30\%$. More detailed results can be found in Appendix \ref{app: add results}. 

\subsubsection{Critical Steps vs. Random Steps}
To further evaluate the efficacy of using a selector LLM for selecting critical steps, we conduct comparisons to randomly selected steps of different ratios. Specifically, we randomly draw 30\%, 20\%, and 10\% steps from the original trajectories and compare them with 30\%, 20\%, and 10\% critical steps for agent tuning.\looseness-1

The results are summarized in Table \ref{tab:abl_random ratio}, with only the average performance across tasks presented due to space constraints. Detailed results are provided in Appendix \ref{app: add results}. Our results indicate that increasing the ratio of random selection improves performance of the fine-tuned LLM agent on both held-in and held-out tasks. However, the random-selection's performance at all the three ratios remains inferior to that achieved on \ours's critical steps, particularly at higher ratios, highlighting the effectiveness of the selector. 
\begin{table}[t]
    \centering
    \small
    \begin{tabular}{llcc}
    \toprule
         \textbf{Data} &\textbf{Steps} & \textbf{held-in} & \textbf{held-out} \\
     \midrule
      Complete &100\%  & 60.52 & 36.18 \\
      \midrule
      Random &30\% & \textbf{59.90} & \textbf{38.04} \\
      Random &20\% & 59.08 & 34.73 \\
      Random &10\% & 58.68 & 29.04 \\
      \midrule
      Critical &30\% & \textbf{65.91} & \textbf{38.36} \\
      Critical &20\% & 60.40 & 36.96 \\
      Critical &10\% & 58.99 & 31.18 \\
      \bottomrule
    \end{tabular}
    \caption{\ours vs. random selection of steps at different ratios (including 100\% as a reference). }
    \label{tab:abl_random ratio}
\end{table}

\subsubsection{Value Function defined Critical Steps}
To further assess the accuracy of using GPT to identify critical steps, we perform experiments to compare datasets selected based on the estimated value function. Following \citep{xiong2024watch}, we estimate the value of expert actions by traversing all actions along the trajectory \(N\) times. Specifically, the agent begins from a specific action in the expert trajectory and performs rollouts from that action \(N\) times. The final rewards from these \(N\) rollouts are summed and averaged to estimate the value of the action. Subsequently, the differences in value between consecutive steps are calculated, and steps with differences exceeding a predefined threshold are identified as critical. We set $N=5$ for this experiment and a detailed methodology is provided in Appendix \ref{app: value estimation}. Notably, the estimated value function is designed for single-task training, so we compare the performance on the BabyAI and Weather tasks for evaluation These environments are chosen because some others randomly generate goals upon reset, making rollouts infeasible. Additionally, environments with long horizons are impractical due to the extensive time and computational resources required. For instance, the rollout of a dataset with 2000 trajectories, each averaging 25 steps, necessitates at least $6.5 \times 10^5$ inferences, which is highly inefficient.

Table \ref{tab: IoU} shows that our method achieves performance comparable to the estimated value function approach. However, with limited rollouts, the estimates are often inaccurate, producing values of 0 at initial steps and 1 at later steps. As a result, only a few steps are identified as critical based on these variations. Despite this limitation, the comparable performance of the two methods suggests the potential for achieving better results with fewer critical steps, which we plan to explore in future work.
\looseness-1

\begin{table}[t]
    \centering
    \small
    \begin{tabular}{llcc}
    \toprule
        Task & Method &Performance \\
    \midrule
        \multirow{2}{*}{BabyAI} & \ours & 78.93        \\
        & Value Func. & 78.58 & \\
        \midrule
        \multirow{2}{*}{Weather} &  \ours &  60.00    \\
        & Value Func. & 60.00 &  \\
        \bottomrule
    \end{tabular}
    \caption{\ours vs. Value function (estimated oracle) on critical step selection for the BabyAI and Weather tasks. 
    The value function follows~\citep{xiong2024watch}. It measures the importance of each step in MDP but it is computationally intractable in practice. 
    This evaluation aims to verify the consistency between LLM-selected critical steps and their values in MDP. \looseness-1}
    \label{tab: IoU}
\end{table}

\subsubsection{Choices of Selector LLMs}
In this section, we evaluate different models for selecting critical steps. Specifically, we use the open-source model Llama3.1-70B-Instruct\footnote{\url{https://huggingface.co/meta-llama/Llama-3.1-70B-Instruct}} as the comparison teacher model to identify critical steps on three tasks Alfworld, BabyAI and Weather. The results in Table \ref{tab:dif teacher} indicate that GPT4o selected dataset can better fine-tune the LLM agents on all three tasks, showing that the selector model's capability significantly impacts the quality of the critical step dataset. The dataset selected by Llama3.1-70B-Instruct fails to accurately identify critical steps, instead including many non-critical steps, which in turn reduces the agents' performance after fine-tuning.
\looseness-1

\begin{table}[t]
    \centering
    \small
    \begin{tabular}{llc}
    \toprule
      Task   & Selector & Performance\\
      \midrule
      \multirow{2}{*}{Alfworld} & GPT-4o & 83.00 \\
      & Llama3.1-70B & 78.50 \\
      \midrule
      \multirow{2}{*}{BabyAI} & GPT-4o & 78.93 \\
      & Llama3.1-70B & 67.23\\
      \midrule
     \multirow{2}{*}{Weather} & GPT-4o & 60.00 \\
      & Llama3.1-70B & 55.00 \\
      \bottomrule
    \end{tabular}
    \caption{\textbf{Ablation study of the selector LLM} for critical step selection on Alfworld, BabyAI, and Weather tasks. Average performance over tasks is reported.}
    \vspace{-4mm}
    \label{tab:dif teacher}
\end{table}

\subsection{Critical Step Verification}

We conduct an additional ablation study to evaluate the improvement in the base model's performance (LLaMA-3.1-8B-Instruct) when execution begins immediately after each identified critical step. In trajectories containing multiple critical steps, the model is reinitialized after each one in an iterative manner, ensuring that all critical steps are utilized throughout the trajectory.

Experimental results on a representative subset of 100 tasks from the BabyAI and Maze environments reveal notable performance improvements when the model is initialized from critical steps, compared to start from the beginning of the task. This indicates that the base model is generally capable of generating all non-critical steps autonomously and that training efforts should prioritize learning from critical steps specifically.

\begin{table}[h!]
\centering
\small
\begin{tabular}{c|c|c}
\hline
Tasks & From Scratch & From Critical Step \\
\hline
BabyAI & 54.8 & 76.4 \\
Maze   & 18.0 & 44.0 \\
\hline
\end{tabular}
\caption{Performance comparison of the model initialized from scratch vs. from critical steps on BabyAI and Maze tasks.}
\end{table}

\section{Related Work}

\subsection{LLM Agents}
LLMs have demonstrated strong capabilities as general-purpose agents, effectively handling complex tasks across diverse environments without task-specific training. Several recent works explore different paradigms to enhance the reasoning, planning, and learning abilities of LLM-based agents. \citep{huang2022languagemodelszeroshotplanners} demonstrate that LLMs can act as zero-shot planners, enabling agents to perform complex tasks directly from natural language instructions. Building on this, Inner Monologue \citep{huang2022innermonologueembodiedreasoning} proposes using internal dialogues for decision-making and planning, further eliminating the need for external training. Similarly, AUTOACT \citep{qiao2024autoactautomaticagentlearning} shows that agents can learn to solve question-answering tasks through self-generated strategies and planning. ReST meets ReAct \citep{aksitov2023restmeetsreactselfimprovement} introduces a self-improvement framework in which agents enhance their multi-step reasoning through iterative feedback. TextGrad \citep{yuksekgonul2024textgradautomaticdifferentiationtext} leverages LLMs for automatic differentiation via textual reasoning, bypassing conventional training pipelines. 

In the realm of interactive and social intelligence, SOTOPIA \citep{wang2024sotopiapiinteractivelearningsocially} emphasizes real-time human-agent communication for acquiring socially intelligent behavior. MMAC-Copilot extends this by employing collaborative chains to improve LLM agents’ performance in multimodal interaction scenarios. \citep{fang2024large} demonstrate the effectiveness of integrating LLMs with external symbolic modules for zero-shot symbolic reasoning in text-based games. To improve agent performance in dynamic environments, MC-DML \citep{shi2025monte} combines LLMs with Monte Carlo planning for text-based decision-making. Lastly, HASARD \citep{tomilin2025hasard} presents a vision-based safe reinforcement learning benchmark that challenges embodied agents with complex egocentric tasks, advancing the evaluation of safety in RL. 


\subsection{Tuning Agents}
Fine-tuning large language models (LLMs) to enhance their performance in agent-related tasks has emerged as a key area of research, with various works proposing innovative frameworks, datasets, and methods. FireAct \citep{chen2023fireact} fine-tunes models using various agent trajectories in the ReAct format, achieving substantial performance boosts. AgentTuning \citep{zeng2023agenttuning} demonstrates the value of instruction tuning with datasets such as AgentInstruct, achieving superior task performance with its AgentLM models. AgentBank \citep{song2024agentbank} improves agent generalization and performance by fine-tuning LLMs with more than 50,000 diverse interaction trajectories. AgentOhana \citep{zhang2024agentohana} standardizes and aggregates agent trajectory datasets, facilitating efficient pipelines and providing state-of-the-art results with its xLAM-v0.1 model. AGENTGYM \citep{xi2024agentgym} enables agent self-evolution through the interactive AGENTEVOL framework, which reduces human supervision through environmental feedback. The IPR framework \citep{xiong2024watch} refines LLM agents with step-level feedback, demonstrating strong results on complex benchmarks. Above all, these works highlight the critical role of fine-tuning in advancing LLM agent capabilities, leveraging diverse data, iterative feedback mechanisms, and innovative frameworks to improve generalization, robustness, and task efficiency.
\looseness-1

\section{Conclusion}
In this work, we propose a novel method, \ours, for fine-tuning LLM agents by focusing on critical steps in expert trajectories. Using an oracle LLM as the selector, we first construct a dataset of critical steps, which highlights the most impactful actions in the trajectories. We then optimize the LLMs by computing and minimizing the loss exclusively on these critical steps. Experimental results demonstrate that by leveraging only 30\% of the steps, our method achieves superior performance in training environments compared to finetuning on 100\% of the trajectories. Moreover, the improved generalization ability of our finetuned agents is also reflected in enhanced performance on held-out tasks. These findings underscore the importance of critical step training in maintaining and enhancing the strengths of LLM agents while significantly reducing training costs. This approach offers a cost-effective solution for fine-tuning LLMs across diverse tasks and environments.
\looseness-1

\section{Limitation}
While our proposed method, \ours, demonstrates significant improvements in efficiency and generalization for fine-tuning LLM agents on multi-environments, some aspects could be further improved. Firstly, the current approach to selecting critical steps depends heavily on powerful closed-source models. It is crucial to explore methods that enable precise critical step selection while minimizing time and computatiional costs. Additionally, the existing selection process primarily focuses on semantic aspects; combining it with other metrics could enhance the accuracy of identifying critical steps and further reduce the tokens needed for fine-tuning. 
\looseness-1

\bibliography{acl_latex}

\begin{thebibliography}{57}
\providecommand{\natexlab}[1]{#1}

\bibitem[{Abdulhai et~al.(2023)Abdulhai, White, Snell, Sun, Hong, Zhai, Xu, and Levine}]{abdulhai2023lmrl}
Marwa Abdulhai, Isadora White, Charlie Snell, Charles Sun, Joey Hong, Yuexiang Zhai, Kelvin Xu, and Sergey Levine. 2023.
\newblock Lmrl gym: Benchmarks for multi-turn reinforcement learning with language models.
\newblock \emph{arXiv preprint arXiv:2311.18232}.

\bibitem[{Aksitov et~al.(2023)Aksitov, Miryoosefi, Li, Li, Babayan, Kopparapu, Fisher, Guo, Prakash, Srinivasan, Zaheer, Yu, and Kumar}]{aksitov2023restmeetsreactselfimprovement}
Renat Aksitov, Sobhan Miryoosefi, Zonglin Li, Daliang Li, Sheila Babayan, Kavya Kopparapu, Zachary Fisher, Ruiqi Guo, Sushant Prakash, Pranesh Srinivasan, Manzil Zaheer, Felix Yu, and Sanjiv Kumar. 2023.
\newblock \href {https://arxiv.org/abs/2312.10003} {Rest meets react: Self-improvement for multi-step reasoning llm agent}.
\newblock \emph{Preprint}, arXiv:2312.10003.

\bibitem[{Anthropic(2024)}]{antro2024claude}
Anthropic. 2024.
\newblock \href {https://www-cdn.anthropic.com/de8ba9b01c9ab7cbabf5c33b80b7bbc618857627/Model_Card_Claude_3.pdf} {The claude 3 model family: Opus, sonnet, haiku}.

\bibitem[{Brown et~al.(2020)Brown, Mann, Ryder, Subbiah, Kaplan, Dhariwal, Neelakantan, Shyam, Sastry, Askell, Agarwal, Herbert-Voss, Krueger, Henighan, Child, Ramesh, Ziegler, Wu, Winter, Hesse, Chen, Sigler, Litwin, Gray, Chess, Clark, Berner, McCandlish, Radford, Sutskever, and Amodei}]{NEURIPS2020_1457c0d6}
Tom Brown, Benjamin Mann, Nick Ryder, Melanie Subbiah, Jared~D Kaplan, Prafulla Dhariwal, Arvind Neelakantan, Pranav Shyam, Girish Sastry, Amanda Askell, Sandhini Agarwal, Ariel Herbert-Voss, Gretchen Krueger, Tom Henighan, Rewon Child, Aditya Ramesh, Daniel Ziegler, Jeffrey Wu, Clemens Winter, Chris Hesse, Mark Chen, Eric Sigler, Mateusz Litwin, Scott Gray, Benjamin Chess, Jack Clark, Christopher Berner, Sam McCandlish, Alec Radford, Ilya Sutskever, and Dario Amodei. 2020.
\newblock \href {https://proceedings.neurips.cc/paper_files/paper/2020/file/1457c0d6bfcb4967418bfb8ac142f64a-Paper.pdf} {Language models are few-shot learners}.
\newblock In \emph{Advances in Neural Information Processing Systems}, volume~33, pages 1877--1901. Curran Associates, Inc.

\bibitem[{Chen et~al.(2023)Chen, Shu, Shareghi, Collier, Narasimhan, and Yao}]{chen2023fireact}
Baian Chen, Chang Shu, Ehsan Shareghi, Nigel Collier, Karthik Narasimhan, and Shunyu Yao. 2023.
\newblock Fireact: Toward language agent fine-tuning.
\newblock \emph{arXiv preprint arXiv:2310.05915}.

\bibitem[{Chen et~al.(2024)Chen, Liu, Wang, Zhang, Liu, Lin, Chen, and Zhao}]{chen2024agent}
Zehui Chen, Kuikun Liu, Qiuchen Wang, Wenwei Zhang, Jiangning Liu, Dahua Lin, Kai Chen, and Feng Zhao. 2024.
\newblock Agent-flan: Designing data and methods of effective agent tuning for large language models.
\newblock \emph{arXiv preprint arXiv:2403.12881}.

\bibitem[{Chevalier-Boisvert et~al.(2018)Chevalier-Boisvert, Bahdanau, Lahlou, Willems, Saharia, Nguyen, and Bengio}]{chevalier2018babyai}
Maxime Chevalier-Boisvert, Dzmitry Bahdanau, Salem Lahlou, Lucas Willems, Chitwan Saharia, Thien~Huu Nguyen, and Yoshua Bengio. 2018.
\newblock Babyai: A platform to study the sample efficiency of grounded language learning.
\newblock \emph{arXiv preprint arXiv:1810.08272}.

\bibitem[{Dubey et~al.(2024)Dubey, Jauhri, Pandey, Kadian, Al-Dahle, Letman, Mathur, Schelten, Yang, Fan et~al.}]{dubey2024llama}
Abhimanyu Dubey, Abhinav Jauhri, Abhinav Pandey, Abhishek Kadian, Ahmad Al-Dahle, Aiesha Letman, Akhil Mathur, Alan Schelten, Amy Yang, Angela Fan, et~al. 2024.
\newblock The llama 3 herd of models.
\newblock \emph{arXiv preprint arXiv:2407.21783}.

\bibitem[{Fang et~al.(2024)Fang, Deng, Zhang, Shi, Chen, Pechenizkiy, and Wang}]{fang2024large}
Meng Fang, Shilong Deng, Yudi Zhang, Zijing Shi, Ling Chen, Mykola Pechenizkiy, and Jun Wang. 2024.
\newblock Large language models are neurosymbolic reasoners.
\newblock In \emph{Proceedings of the AAAI conference on artificial intelligence}, volume~38, pages 17985--17993.

\bibitem[{Gemma~Team et~al.(2024)Gemma~Team, Hardin, Dadashi, Bhupatiraju, Sifre, Rivière, Kale, Love, Tafti, Hussenot, and et~al.}]{gemma_2024}
Thomas~Mesnard Gemma~Team, Cassidy Hardin, Robert Dadashi, Surya Bhupatiraju, Laurent Sifre, Morgane Rivière, Mihir~Sanjay Kale, Juliette Love, Pouya Tafti, Léonard Hussenot, and et~al. 2024.
\newblock \href {https://doi.org/10.34740/KAGGLE/M/3301} {Gemma}.

\bibitem[{Ghosh et~al.(2024)Ghosh, Evuru, Kumar, Aneja, Jin, Duraiswami, Manocha et~al.}]{ghosh2024closer}
Sreyan Ghosh, Chandra Kiran~Reddy Evuru, Sonal Kumar, Deepali Aneja, Zeyu Jin, Ramani Duraiswami, Dinesh Manocha, et~al. 2024.
\newblock A closer look at the limitations of instruction tuning.
\newblock \emph{arXiv preprint arXiv:2402.05119}.

\bibitem[{Huang et~al.(2022{\natexlab{a}})Huang, Abbeel, Pathak, and Mordatch}]{huang2022languagemodelszeroshotplanners}
Wenlong Huang, Pieter Abbeel, Deepak Pathak, and Igor Mordatch. 2022{\natexlab{a}}.
\newblock \href {https://arxiv.org/abs/2201.07207} {Language models as zero-shot planners: Extracting actionable knowledge for embodied agents}.
\newblock \emph{Preprint}, arXiv:2201.07207.

\bibitem[{Huang et~al.(2022{\natexlab{b}})Huang, Xia, Xiao, Chan, Liang, Florence, Zeng, Tompson, Mordatch, Chebotar, Sermanet, Brown, Jackson, Luu, Levine, Hausman, and Ichter}]{huang2022innermonologueembodiedreasoning}
Wenlong Huang, Fei Xia, Ted Xiao, Harris Chan, Jacky Liang, Pete Florence, Andy Zeng, Jonathan Tompson, Igor Mordatch, Yevgen Chebotar, Pierre Sermanet, Noah Brown, Tomas Jackson, Linda Luu, Sergey Levine, Karol Hausman, and Brian Ichter. 2022{\natexlab{b}}.
\newblock \href {https://arxiv.org/abs/2207.05608} {Inner monologue: Embodied reasoning through planning with language models}.
\newblock \emph{Preprint}, arXiv:2207.05608.

\bibitem[{Jiang et~al.(2023)Jiang, Sablayrolles, Mensch, Bamford, Chaplot, Casas, Bressand, Lengyel, Lample, Saulnier et~al.}]{jiang2023mistral}
Albert~Q Jiang, Alexandre Sablayrolles, Arthur Mensch, Chris Bamford, Devendra~Singh Chaplot, Diego de~las Casas, Florian Bressand, Gianna Lengyel, Guillaume Lample, Lucile Saulnier, et~al. 2023.
\newblock Mistral 7b.
\newblock \emph{arXiv preprint arXiv:2310.06825}.

\bibitem[{Kingma(2014)}]{kingma2014adam}
Diederik~P Kingma. 2014.
\newblock Adam: A method for stochastic optimization.
\newblock \emph{arXiv preprint arXiv:1412.6980}.

\bibitem[{Kojima et~al.(2022)Kojima, Gu, Reid, Matsuo, and Iwasawa}]{kojima2022large}
Takeshi Kojima, Shixiang~Shane Gu, Machel Reid, Yutaka Matsuo, and Yusuke Iwasawa. 2022.
\newblock Large language models are zero-shot reasoners.
\newblock \emph{Advances in neural information processing systems}, 35:22199--22213.

\bibitem[{Li et~al.(2024{\natexlab{a}})Li, Chen, Chen, He, Gu, and Zhou}]{li-etal-2024-selective}
Ming Li, Lichang Chen, Jiuhai Chen, Shwai He, Jiuxiang Gu, and Tianyi Zhou. 2024{\natexlab{a}}.
\newblock \href {https://aclanthology.org/2024.findings-acl.958} {Selective reflection-tuning: Student-selected data recycling for {LLM} instruction-tuning}.
\newblock In \emph{Findings of the Association for Computational Linguistics ACL 2024}, pages 16189--16211, Bangkok, Thailand and virtual meeting. Association for Computational Linguistics.

\bibitem[{Li et~al.(2023)Li, Chen, Chen, He, and Zhou}]{li2023reflectiontuning}
Ming Li, Lichang Chen, Jiuhai Chen, Shwai He, and Tianyi Zhou. 2023.
\newblock \href {https://openreview.net/forum?id=xaqoZZqkPU} {Reflection-tuning: Recycling data for better instruction-tuning}.
\newblock In \emph{NeurIPS 2023 Workshop on Instruction Tuning and Instruction Following}.

\bibitem[{Li et~al.(2024{\natexlab{b}})Li, Chen, Wang, Zhao, Liang, Hou, Liu, and Zhou}]{li2024mosaic}
Ming Li, Pei Chen, Chenguang Wang, Hongyu Zhao, Yijun Liang, Yupeng Hou, Fuxiao Liu, and Tianyi Zhou. 2024{\natexlab{b}}.
\newblock Mosaic-it: Free compositional data augmentation improves instruction tuning.
\newblock \emph{arXiv preprint arXiv:2405.13326}.

\bibitem[{Li et~al.(2024{\natexlab{c}})Li, Zhang, He, Li, Zhao, Wang, Cheng, and Zhou}]{li-etal-2024-superfiltering}
Ming Li, Yong Zhang, Shwai He, Zhitao Li, Hongyu Zhao, Jianzong Wang, Ning Cheng, and Tianyi Zhou. 2024{\natexlab{c}}.
\newblock \href {https://aclanthology.org/2024.acl-long.769} {Superfiltering: Weak-to-strong data filtering for fast instruction-tuning}.
\newblock In \emph{Proceedings of the 62nd Annual Meeting of the Association for Computational Linguistics (Volume 1: Long Papers)}, pages 14255--14273, Bangkok, Thailand. Association for Computational Linguistics.

\bibitem[{Li et~al.(2024{\natexlab{d}})Li, Zhang, Li, Chen, Chen, Cheng, Wang, Zhou, and Xiao}]{li-etal-2024-quantity}
Ming Li, Yong Zhang, Zhitao Li, Jiuhai Chen, Lichang Chen, Ning Cheng, Jianzong Wang, Tianyi Zhou, and Jing Xiao. 2024{\natexlab{d}}.
\newblock \href {https://aclanthology.org/2024.naacl-long.421} {From quantity to quality: Boosting {LLM} performance with self-guided data selection for instruction tuning}.
\newblock In \emph{Proceedings of the 2024 Conference of the North American Chapter of the Association for Computational Linguistics: Human Language Technologies (Volume 1: Long Papers)}, pages 7595--7628, Mexico City, Mexico. Association for Computational Linguistics.

\bibitem[{Liu et~al.(2024)Liu, Feng, Wang, Wang, Liu, Zhao, Dengr, Ruan, Dai, Guo et~al.}]{liu2024deepseek}
Aixin Liu, Bei Feng, Bin Wang, Bingxuan Wang, Bo~Liu, Chenggang Zhao, Chengqi Dengr, Chong Ruan, Damai Dai, Daya Guo, et~al. 2024.
\newblock Deepseek-v2: A strong, economical, and efficient mixture-of-experts language model.
\newblock \emph{arXiv preprint arXiv:2405.04434}.

\bibitem[{Ma et~al.(2024)Ma, Zhang, Zhu, Yang, Yang, Jin, Lan, Kong, and He}]{ma2024agentboard}
Chang Ma, Junlei Zhang, Zhihao Zhu, Cheng Yang, Yujiu Yang, Yaohui Jin, Zhenzhong Lan, Lingpeng Kong, and Junxian He. 2024.
\newblock Agentboard: An analytical evaluation board of multi-turn llm agents.
\newblock \emph{arXiv preprint arXiv:2401.13178}.

\bibitem[{Maes(1995)}]{maes1995agents}
Pattie Maes. 1995.
\newblock Agents that reduce work and information overload.
\newblock In \emph{Readings in human--computer interaction}, pages 811--821. Elsevier.

\bibitem[{Ouyang et~al.(2022)Ouyang, Wu, Jiang, Almeida, Wainwright, Mishkin, Zhang, Agarwal, Slama, Ray et~al.}]{ouyang2022training}
Long Ouyang, Jeffrey Wu, Xu~Jiang, Diogo Almeida, Carroll Wainwright, Pamela Mishkin, Chong Zhang, Sandhini Agarwal, Katarina Slama, Alex Ray, et~al. 2022.
\newblock Training language models to follow instructions with human feedback.
\newblock \emph{Advances in neural information processing systems}, 35:27730--27744.

\bibitem[{Paszke et~al.(2019)Paszke, Gross, Massa, Lerer, Bradbury, Chanan, Killeen, Lin, Gimelshein, Antiga et~al.}]{paszke2019pytorch}
Adam Paszke, Sam Gross, Francisco Massa, Adam Lerer, James Bradbury, Gregory Chanan, Trevor Killeen, Zeming Lin, Natalia Gimelshein, Luca Antiga, et~al. 2019.
\newblock Pytorch: An imperative style, high-performance deep learning library.
\newblock \emph{Advances in neural information processing systems}, 32.

\bibitem[{Patil et~al.(2023)Patil, Zhang, Wang, and Gonzalez}]{patil2023gorilla}
Shishir~G Patil, Tianjun Zhang, Xin Wang, and Joseph~E Gonzalez. 2023.
\newblock Gorilla: Large language model connected with massive apis.
\newblock \emph{arXiv preprint arXiv:2305.15334}.

\bibitem[{Prasad et~al.(2023)Prasad, Koller, Hartmann, Clark, Sabharwal, Bansal, and Khot}]{prasad2023adapt}
Archiki Prasad, Alexander Koller, Mareike Hartmann, Peter Clark, Ashish Sabharwal, Mohit Bansal, and Tushar Khot. 2023.
\newblock Adapt: As-needed decomposition and planning with language models.
\newblock \emph{arXiv preprint arXiv:2311.05772}.

\bibitem[{Qiao et~al.(2024)Qiao, Zhang, Fang, Luo, Zhou, Jiang, Lv, and Chen}]{qiao2024autoactautomaticagentlearning}
Shuofei Qiao, Ningyu Zhang, Runnan Fang, Yujie Luo, Wangchunshu Zhou, Yuchen~Eleanor Jiang, Chengfei Lv, and Huajun Chen. 2024.
\newblock \href {https://arxiv.org/abs/2401.05268} {Autoact: Automatic agent learning from scratch for qa via self-planning}.
\newblock \emph{Preprint}, arXiv:2401.05268.

\bibitem[{Qin et~al.(2023)Qin, Liang, Ye, Zhu, Yan, Lu, Lin, Cong, Tang, Qian et~al.}]{qin2023toolllm}
Yujia Qin, Shihao Liang, Yining Ye, Kunlun Zhu, Lan Yan, Yaxi Lu, Yankai Lin, Xin Cong, Xiangru Tang, Bill Qian, et~al. 2023.
\newblock Toolllm: Facilitating large language models to master 16000+ real-world apis.
\newblock \emph{arXiv preprint arXiv:2307.16789}.

\bibitem[{Shi et~al.(2025)Shi, Fang, and Chen}]{shi2025monte}
Zijing Shi, Meng Fang, and Ling Chen. 2025.
\newblock Monte carlo planning with large language model for text-based game agents.
\newblock \emph{arXiv preprint arXiv:2504.16855}.

\bibitem[{Shridhar et~al.(2020)Shridhar, Yuan, C{\^o}t{\'e}, Bisk, Trischler, and Hausknecht}]{shridhar2020alfworld}
Mohit Shridhar, Xingdi Yuan, Marc-Alexandre C{\^o}t{\'e}, Yonatan Bisk, Adam Trischler, and Matthew Hausknecht. 2020.
\newblock Alfworld: Aligning text and embodied environments for interactive learning.
\newblock \emph{arXiv preprint arXiv:2010.03768}.

\bibitem[{Silver et~al.(2017)Silver, Schrittwieser, Simonyan, Antonoglou, Huang, Guez, Hubert, Baker, Lai, Bolton et~al.}]{silver2017mastering}
David Silver, Julian Schrittwieser, Karen Simonyan, Ioannis Antonoglou, Aja Huang, Arthur Guez, Thomas Hubert, Lucas Baker, Matthew Lai, Adrian Bolton, et~al. 2017.
\newblock Mastering the game of go without human knowledge.
\newblock \emph{nature}, 550(7676):354--359.

\bibitem[{Singh et~al.(2024)Singh, Co-Reyes, Agarwal, Anand, Patil, Garcia, Liu, Harrison, Lee, Xu, Parisi, Kumar, Alemi, Rizkowsky, Nova, Adlam, Bohnet, Elsayed, Sedghi, Mordatch, Simpson, Gur, Snoek, Pennington, Hron, Kenealy, Swersky, Mahajan, Culp, Xiao, Bileschi, Constant, Novak, Liu, Warkentin, Qian, Bansal, Dyer, Neyshabur, Sohl-Dickstein, and Fiedel}]{singh2024humandatascalingselftraining}
Avi Singh, John~D. Co-Reyes, Rishabh Agarwal, Ankesh Anand, Piyush Patil, Xavier Garcia, Peter~J. Liu, James Harrison, Jaehoon Lee, Kelvin Xu, Aaron Parisi, Abhishek Kumar, Alex Alemi, Alex Rizkowsky, Azade Nova, Ben Adlam, Bernd Bohnet, Gamaleldin Elsayed, Hanie Sedghi, Igor Mordatch, Isabelle Simpson, Izzeddin Gur, Jasper Snoek, Jeffrey Pennington, Jiri Hron, Kathleen Kenealy, Kevin Swersky, Kshiteej Mahajan, Laura Culp, Lechao Xiao, Maxwell~L. Bileschi, Noah Constant, Roman Novak, Rosanne Liu, Tris Warkentin, Yundi Qian, Yamini Bansal, Ethan Dyer, Behnam Neyshabur, Jascha Sohl-Dickstein, and Noah Fiedel. 2024.
\newblock \href {https://arxiv.org/abs/2312.06585} {Beyond human data: Scaling self-training for problem-solving with language models}.
\newblock \emph{Preprint}, arXiv:2312.06585.

\bibitem[{Song et~al.(2024)Song, Xiong, Zhao, Zhu, Wu, Wang, Li, Peng, and Li}]{song2024agentbank}
Yifan Song, Weimin Xiong, Xiutian Zhao, Dawei Zhu, Wenhao Wu, Ke~Wang, Cheng Li, Wei Peng, and Sujian Li. 2024.
\newblock Agentbank: Towards generalized llm agents via fine-tuning on 50000+ interaction trajectories.
\newblock \emph{arXiv preprint arXiv:2410.07706}.

\bibitem[{Sutton(2018)}]{sutton2018reinforcement}
Richard~S Sutton. 2018.
\newblock Reinforcement learning: An introduction.
\newblock \emph{A Bradford Book}.

\bibitem[{Team(2024{\natexlab{a}})}]{openai2024gpt4technicalreport}
OpenAI Team. 2024{\natexlab{a}}.
\newblock \href {https://arxiv.org/abs/2303.08774} {Gpt-4 technical report}.
\newblock \emph{Preprint}, arXiv:2303.08774.

\bibitem[{Team(2024{\natexlab{b}})}]{phi3}
Phi-3 Team. 2024{\natexlab{b}}.
\newblock Phi-3 technical report: A highly capable language model locally on your phone.
\newblock \emph{arXiv preprint arXiv:2407.10671}.

\bibitem[{Team(2024{\natexlab{c}})}]{qwen2}
Qwen2 Team. 2024{\natexlab{c}}.
\newblock Qwen2 technical report.
\newblock \emph{arXiv preprint arXiv:2407.10671}.

\bibitem[{Tomilin et~al.(2025)Tomilin, Fang, and Pechenizkiy}]{tomilin2025hasard}
Tristan Tomilin, Meng Fang, and Mykola Pechenizkiy. 2025.
\newblock Hasard: A benchmark for vision-based safe reinforcement learning in embodied agents.
\newblock \emph{arXiv preprint arXiv:2503.08241}.

\bibitem[{Touvron et~al.(2023{\natexlab{a}})Touvron, Lavril, Izacard, Martinet, Lachaux, Lacroix, Rozière, Goyal, Hambro, Azhar, Rodriguez, Joulin, Grave, and Lample}]{touvron2023llama}
Hugo Touvron, Thibaut Lavril, Gautier Izacard, Xavier Martinet, Marie-Anne Lachaux, Timothée Lacroix, Baptiste Rozière, Naman Goyal, Eric Hambro, Faisal Azhar, Aurelien Rodriguez, Armand Joulin, Edouard Grave, and Guillaume Lample. 2023{\natexlab{a}}.
\newblock \href {https://arxiv.org/abs/2302.13971} {Llama: Open and efficient foundation language models}.
\newblock \emph{Preprint}, arXiv:2302.13971.

\bibitem[{Touvron et~al.(2023{\natexlab{b}})Touvron, Martin, Stone, Albert, Almahairi, Babaei, Bashlykov, Batra, Bhargava, Bhosale, Bikel, Blecher, Ferrer, Chen, Cucurull, Esiobu, Fernandes, Fu, Fu, Fuller, Gao, Goswami, Goyal, Hartshorn, Hosseini, Hou, Inan, Kardas, Kerkez, Khabsa, Kloumann, Korenev, Koura, Lachaux, Lavril, Lee, Liskovich, Lu, Mao, Martinet, Mihaylov, Mishra, Molybog, Nie, Poulton, Reizenstein, Rungta, Saladi, Schelten, Silva, Smith, Subramanian, Tan, Tang, Taylor, Williams, Kuan, Xu, Yan, Zarov, Zhang, Fan, Kambadur, Narang, Rodriguez, Stojnic, Edunov, and Scialom}]{touvron2023llama2}
Hugo Touvron, Louis Martin, Kevin Stone, Peter Albert, Amjad Almahairi, Yasmine Babaei, Nikolay Bashlykov, Soumya Batra, Prajjwal Bhargava, Shruti Bhosale, Dan Bikel, Lukas Blecher, Cristian~Canton Ferrer, Moya Chen, Guillem Cucurull, David Esiobu, Jude Fernandes, Jeremy Fu, Wenyin Fu, Brian Fuller, Cynthia Gao, Vedanuj Goswami, Naman Goyal, Anthony Hartshorn, Saghar Hosseini, Rui Hou, Hakan Inan, Marcin Kardas, Viktor Kerkez, Madian Khabsa, Isabel Kloumann, Artem Korenev, Punit~Singh Koura, Marie-Anne Lachaux, Thibaut Lavril, Jenya Lee, Diana Liskovich, Yinghai Lu, Yuning Mao, Xavier Martinet, Todor Mihaylov, Pushkar Mishra, Igor Molybog, Yixin Nie, Andrew Poulton, Jeremy Reizenstein, Rashi Rungta, Kalyan Saladi, Alan Schelten, Ruan Silva, Eric~Michael Smith, Ranjan Subramanian, Xiaoqing~Ellen Tan, Binh Tang, Ross Taylor, Adina Williams, Jian~Xiang Kuan, Puxin Xu, Zheng Yan, Iliyan Zarov, Yuchen Zhang, Angela Fan, Melanie Kambadur, Sharan Narang, Aurelien Rodriguez, Robert Stojnic, Sergey Edunov, and Thomas
  Scialom. 2023{\natexlab{b}}.
\newblock \href {https://arxiv.org/abs/2307.09288} {Llama 2: Open foundation and fine-tuned chat models}.
\newblock \emph{Preprint}, arXiv:2307.09288.

\bibitem[{Vinyals et~al.(2019)Vinyals, Babuschkin, Czarnecki, Mathieu, Dudzik, Chung, Choi, Powell, Ewalds, Georgiev et~al.}]{vinyals2019grandmaster}
Oriol Vinyals, Igor Babuschkin, Wojciech~M Czarnecki, Micha{\"e}l Mathieu, Andrew Dudzik, Junyoung Chung, David~H Choi, Richard Powell, Timo Ewalds, Petko Georgiev, et~al. 2019.
\newblock Grandmaster level in starcraft ii using multi-agent reinforcement learning.
\newblock \emph{nature}, 575(7782):350--354.

\bibitem[{Wang et~al.(2024{\natexlab{a}})Wang, Ma, Feng, Zhang, Yang, Zhang, Chen, Tang, Chen, Lin et~al.}]{wang2024survey}
Lei Wang, Chen Ma, Xueyang Feng, Zeyu Zhang, Hao Yang, Jingsen Zhang, Zhiyuan Chen, Jiakai Tang, Xu~Chen, Yankai Lin, et~al. 2024{\natexlab{a}}.
\newblock A survey on large language model based autonomous agents.
\newblock \emph{Frontiers of Computer Science}, 18(6):186345.

\bibitem[{Wang et~al.(2024{\natexlab{b}})Wang, Yu, Zhang, Qi, Sap, Neubig, Bisk, and Zhu}]{wang2024sotopiapiinteractivelearningsocially}
Ruiyi Wang, Haofei Yu, Wenxin Zhang, Zhengyang Qi, Maarten Sap, Graham Neubig, Yonatan Bisk, and Hao Zhu. 2024{\natexlab{b}}.
\newblock \href {https://arxiv.org/abs/2403.08715} {Sotopia-$\pi$: Interactive learning of socially intelligent language agents}.
\newblock \emph{Preprint}, arXiv:2403.08715.

\bibitem[{Wang et~al.(2022)Wang, Jansen, C{\^o}t{\'e}, and Ammanabrolu}]{wang2022scienceworld}
Ruoyao Wang, Peter Jansen, Marc-Alexandre C{\^o}t{\'e}, and Prithviraj Ammanabrolu. 2022.
\newblock Scienceworld: Is your agent smarter than a 5th grader?
\newblock \emph{arXiv preprint arXiv:2203.07540}.

\bibitem[{Wei et~al.(2022)Wei, Wang, Schuurmans, Bosma, Xia, Chi, Le, Zhou et~al.}]{wei2022chain}
Jason Wei, Xuezhi Wang, Dale Schuurmans, Maarten Bosma, Fei Xia, Ed~Chi, Quoc~V Le, Denny Zhou, et~al. 2022.
\newblock Chain-of-thought prompting elicits reasoning in large language models.
\newblock \emph{Advances in neural information processing systems}, 35:24824--24837.

\bibitem[{Wooldridge and Jennings(1995)}]{wooldridge1995intelligent}
Michael Wooldridge and Nicholas~R Jennings. 1995.
\newblock Intelligent agents: Theory and practice.
\newblock \emph{The knowledge engineering review}, 10(2):115--152.

\bibitem[{Xi et~al.(2024)Xi, Ding, Chen, Hong, Guo, Wang, Yang, Liao, Guo, He et~al.}]{xi2024agentgym}
Zhiheng Xi, Yiwen Ding, Wenxiang Chen, Boyang Hong, Honglin Guo, Junzhe Wang, Dingwen Yang, Chenyang Liao, Xin Guo, Wei He, et~al. 2024.
\newblock Agentgym: Evolving large language model-based agents across diverse environments.
\newblock \emph{arXiv preprint arXiv:2406.04151}.

\bibitem[{Xiong et~al.(2024)Xiong, Song, Zhao, Wu, Wang, Wang, Li, Peng, and Li}]{xiong2024watch}
Weimin Xiong, Yifan Song, Xiutian Zhao, Wenhao Wu, Xun Wang, Ke~Wang, Cheng Li, Wei Peng, and Sujian Li. 2024.
\newblock Watch every step! llm agent learning via iterative step-level process refinement.
\newblock \emph{arXiv preprint arXiv:2406.11176}.

\bibitem[{Xu et~al.(2024)Xu, Li, Tao, Shen, Cheng, Li, Xu, Tao, and Zhou}]{Xu2024ASO}
Xiaohan Xu, Ming Li, Chongyang Tao, Tao Shen, Reynold Cheng, Jinyang Li, Can Xu, Dacheng Tao, and Tianyi Zhou. 2024.
\newblock \href {https://api.semanticscholar.org/CorpusID:267760021} {A survey on knowledge distillation of large language models}.
\newblock \emph{ArXiv}, abs/2402.13116.

\bibitem[{Yao et~al.(2022{\natexlab{a}})Yao, Chen, Yang, and Narasimhan}]{yao2022webshop}
Shunyu Yao, Howard Chen, John Yang, and Karthik Narasimhan. 2022{\natexlab{a}}.
\newblock Webshop: Towards scalable real-world web interaction with grounded language agents.
\newblock \emph{Advances in Neural Information Processing Systems}, 35:20744--20757.

\bibitem[{Yao et~al.(2022{\natexlab{b}})Yao, Zhao, Yu, Du, Shafran, Narasimhan, and Cao}]{yaoreact}
Shunyu Yao, Jeffrey Zhao, Dian Yu, Nan Du, Izhak Shafran, Karthik~R Narasimhan, and Yuan Cao. 2022{\natexlab{b}}.
\newblock React: Synergizing reasoning and acting in language models.
\newblock In \emph{The Eleventh International Conference on Learning Representations}.

\bibitem[{Yuksekgonul et~al.(2024)Yuksekgonul, Bianchi, Boen, Liu, Huang, Guestrin, and Zou}]{yuksekgonul2024textgradautomaticdifferentiationtext}
Mert Yuksekgonul, Federico Bianchi, Joseph Boen, Sheng Liu, Zhi Huang, Carlos Guestrin, and James Zou. 2024.
\newblock \href {https://arxiv.org/abs/2406.07496} {Textgrad: Automatic "differentiation" via text}.
\newblock \emph{Preprint}, arXiv:2406.07496.

\bibitem[{Zeng et~al.(2023)Zeng, Liu, Lu, Wang, Liu, Dong, and Tang}]{zeng2023agenttuning}
Aohan Zeng, Mingdao Liu, Rui Lu, Bowen Wang, Xiao Liu, Yuxiao Dong, and Jie Tang. 2023.
\newblock Agenttuning: Enabling generalized agent abilities for llms.
\newblock \emph{arXiv preprint arXiv:2310.12823}.

\bibitem[{Zhang et~al.(2024)Zhang, Lan, Murthy, Liu, Yao, Tan, Hoang, Yang, Feng, Liu et~al.}]{zhang2024agentohana}
Jianguo Zhang, Tian Lan, Rithesh Murthy, Zhiwei Liu, Weiran Yao, Juntao Tan, Thai Hoang, Liangwei Yang, Yihao Feng, Zuxin Liu, et~al. 2024.
\newblock Agentohana: Design unified data and training pipeline for effective agent learning.
\newblock \emph{arXiv preprint arXiv:2402.15506}.

\bibitem[{Zhao et~al.(2023)Zhao, Zhou, Li, Tang, Wang, Hou, Min, Zhang, Zhang, Dong, Du, Yang, Chen, Chen, Jiang, Ren, Li, Tang, Liu, Liu, Nie, and Wen}]{zhao2023survey}
Wayne~Xin Zhao, Kun Zhou, Junyi Li, Tianyi Tang, Xiaolei Wang, Yupeng Hou, Yingqian Min, Beichen Zhang, Junjie Zhang, Zican Dong, Yifan Du, Chen Yang, Yushuo Chen, Zhipeng Chen, Jinhao Jiang, Ruiyang Ren, Yifan Li, Xinyu Tang, Zikang Liu, Peiyu Liu, Jian-Yun Nie, and Ji-Rong Wen. 2023.
\newblock \href {https://arxiv.org/abs/2303.18223} {A survey of large language models}.
\newblock \emph{Preprint}, arXiv:2303.18223.

\end{thebibliography}
\newpage
\appendix

\label{sec:appendix}
\section{Environment Details}\label{env_detail}
\textbf{ALFWorld} \citep{shridhar2020alfworld}: ALFWorld is a simulated household environment developed upon the TextWorld framework, designed to require agents to perform tasks that involve spatial navigation and the application of common-sense knowledge. Within this environment, the action space includes interactions such as object manipulation (e.g., lifting, repositioning), environmental inspection, and the operational use of furniture. Agent actions are executed according to predefined logical rules, with the system generating contextually appropriate feedback. The principal evaluation criterion is the success rate, with each task sequence capped at 30 iterations to balance efficiency and feasibility. For the AGENTTRAJ-L dataset, a total of 2,420 trajectories were aggregated, consisting of 1,920 instances generated by state-of-the-art computational models and 500 human-annotated examples to ensure diversity and grounding in real-world reasoning.

\textbf{BabyAI} \citep{chevalier2018babyai}: BabyAI is a grid-world simulation platform featuring 40 instruction-following tasks that necessitate agent-object interaction. Agents are constrained to a 7x7 observational grid, permitting object manipulation only within immediate proximity. The original framework employs visual observations and primitive actions (e.g., "move forward," "turn left"), while an adapted version substitutes visual inputs with textual instructions and integrates high-level action commands (e.g., "pick up green key 1," "go through blue locked door 2"), thereby broadening the action space. A step-discounted reward is assigned upon successful task completion, with unsuccessful attempts yielding zero reward. For AGENTTRAJ-L, 810 trajectories were generated using state-of-the-art models. Performance is evaluated via task-specific rewards, capped at 20 rounds per task.

\textbf{MAZE} \citep{abdulhai2023lmrl}: Maze is a textual environment where agents possess positional awareness, including their current location, target destination, and adjacent obstructions. Agents select movement in one of four directional axes (up, down, left, right) per step, with each action incurring a penalty of -1 until goal attainment. AGENTTRAJ-L comprises 215 trajectories. Performance is evaluated via success rate, with task sequences capped at 15 iterations.

\textbf{Movie} \citep{ma2024agentboard}: Movie is a structured environment enabling LLM agents to leverage a specialized tool for accessing film-related metadata (e.g., cinematic details, personnel, production entities). The framework offers 16 discrete actions to execute task-specific objectives. It integrates The Movie Database (TMDB) API, incorporating its dataset and operational functions to facilitate agent interactions. Agents obtain a binary reward (1 for correct output alignment with reference solutions, 0 otherwise). AGENTTRAJ-L comprises 215 annotated trajectories, with 20 questions reserved for evaluation and the remainder allocated to training. Success rate serves as the primary performance metric, with task sequences limited to 12 iterations.

\textbf{SciWorld} \citep{wang2022scienceworld}: SciWorld functions as an empirical assessment framework for evaluating scientific reasoning capabilities in an interactive textual environment, structured to reflect elementary-level curricular standards. The platform encompasses 30 diverse task categories, operationalizing activities such as tool-based measurement and mechanical experimentation. It employs a domain-specific action space, with a simulator generating contextual feedback on action outcomes. The AGENTTRAJ-L dataset includes 2,120 trajectories synthesized for evaluation. Task performance is measured via reward accumulation, with each trial restricted to a maximum of 30 iterations.

\textbf{TextCraft} \citep{prasad2023adapt}: TextCraft is a text-based environment modeled after WordCraft, designed for simulating item crafting through a hierarchical framework of 544 nodes aligned with Minecraft’s recipe hierarchy. Each task specifies a target item and a sequence of compositional crafting actions (e.g., “craft <item> using <ingredients>,” “get <item>,” “inventory”), with complexity ranging from 1 to 4 procedural steps. The environment provides real-time feedback on action validity and execution states, enabling agents to directly acquire non-craftable items. A reward of 1 is granted exclusively upon successful synthesis of the target item. For evaluation, 100 tasks are partitioned from a broader training set, with AGENTTRAJ-L comprising 374 trajectories (299 generated by state-of-the-art models, 75 human-annotated). Success rate is the primary metric, capped at 20 rounds per task for empirical validation.

\textbf{TODOList} \citep{ma2024agentboard}: Todolist is a structured environment enabling LLM agents to manage personal agenda data through a task-oriented interface with 11 discrete operational commands. The tool integrates the TodoList API to operationalize task management functionalities. Agents receive binary reward allocation (1 for congruence with reference outputs, 0 otherwise). AGENTTRAJ-L comprises 135 trajectories, with 20 task instances reserved for evaluation and the remainder allocated to training. Performance is assessed via success rate, constrained to a maximum of 15 iterations per task.

\textbf{Weather} \citep{ma2024agentboard}: Weather is a structured environment enabling LLM agents to retrieve meteorological parameters (e.g., temperature, precipitation, air quality) across spatiotemporal contexts via a specialized tool. The framework offers 18 discrete operational commands, integrating the Open-Meteo API through Python-based implementation to enable data querying functionalities. Agents receive a binary reward (1 for output alignment with reference solutions, 0 otherwise). AGENTTRAJ-L includes 311 annotated trajectories, with 20 queries reserved for evaluation and the remainder allocated to training. Performance is assessed via success rate, constrained to a maximum of 10 iterations per task.

\textbf{WebShop} \citep{yao2022webshop}: Webshop is a simulated e-commerce platform where agents execute product procurement tasks adhering to predefined criteria through interface interactions (button-based navigation) or text-based search functionality. The environment integrates 12,000 structured instructions and leverages over one million real-world Amazon product listings, with 6,910 instructions selected for task execution. AGENTTRAJ-L includes 3,930 annotated trajectories. Performance is quantified via success rate, with task sequences limited to 10 rounds to balance efficiency and practical applicability.

\textbf{Wordle} \citep{abdulhai2023lmrl}: Wordle is a lexical reasoning assessment framework where agents deduce a target word from a constrained five-letter vocabulary. The environment operationalizes character-level feedback after each guess, indicating positional accuracy and presence of characters. Agents accumulate a step penalty of -1 until successful identification or attempt exhaustion. AGENTTRAJ-L contains 955 trajectories. Performance is quantified via success rate, with trials capped at 8 rounds for empirical validation.

\textbf{Academia} \citep{ma2024agentboard}: Academia is a structured environment enabling LLM agents to access computer science research resources (e.g., publications, author metadata) via seven discrete operational commands. The tool integrates the Citation Network Dataset to implement core functionalities. Agents receive binary rewards (1 for output alignment with reference solutions, 0 otherwise). Performance is assessed via success rate across 20 evaluation tasks, with trials capped at 12 rounds per task.

\textbf{Sheet} \citep{ma2024agentboard}: Sheet is a structured environment enabling LLM agents to manipulate spreadsheet data via 20 discrete operational commands, leveraging the Google Sheets API. Reward is determined by structural and content congruence between the agent-modified spreadsheet and a reference template, quantified on a 0–1 scale. Evaluation employs 20 predefined tasks, with performance measured via reward accumulation and trials capped at 15 rounds per task.

\textbf{Jericho} \citep{ma2024agentboard}: Jericho is a text-based game framework designed to evaluate agents’ capacity for interactive exploration and dynamic world modeling within fictional narratives. These tasks demand agents to infer contextual rules (e.g., magical systems) through iterative interaction rather than pre-existing commonsense knowledge. To address operational feasibility given LLM agents’ context window constraints, original game objectives (often requiring 50–300 steps) are restructured into modular subtasks achievable within 15 steps. For example, the \texttt{zork1} dungeon exploration is redefined as “locate the secret passage entrance in the living room,” reducing the sequence to 8 steps. This adaptation preserves core exploratory challenges while aligning task complexity with computational tractability.

\textbf{PDDL} \citep{ma2024agentboard}: The Planning Domain Definition Language (PDDL) serves as a framework for evaluating strategic reasoning in symbolic planning tasks, with four benchmark domains—Gripper, Barman, Blocksworld, and Tyreworld—designed to test multi-step action sequencing under efficiency constraints. Agents must navigate domain-specific objectives (e.g., transporting objects, mixology, block stacking, tire installation) by optimizing action sequences to minimize redundant operations. For instance, in the Barman domain, agents strategically allocate containers to reduce cleaning cycles during cocktail preparation. To enable natural language interaction, symbolic PDDL predicates (e.g., \texttt{ontable(shaker1)}) and actions (e.g., \texttt{clean-shaker}) are translated into textual observations (e.g., “Shaker1 is on the table”) and instructions (e.g., “Clean shaker1 with hand1 while hand2 is empty”). Each domain features 10–20 curated multi-round problems, with progress measured via a normalized matching score that quantifies alignment between the current state and the goal state. Full task completion (100\% progress) requires satisfying all goal conditions, such as hierarchical block placement in Blocksworld. This adaptation bridges symbolic planning with language-agent interoperability while preserving strategic complexity.

\section{More Implementation Details}\label{train_conf}
\subsection{Training Configuration}
We use the Adam optimizer \citep{kingma2014adam}, with a learning rate of 2e-5 and a cosine scheduler for the agent fine-tuning. The models are trained with 3 epoches and a warmup rate 0.03. The batch size is 128 and the max length of 8192. All experiments are conducted on 4 NVIDIA A100 80G GPUs and fine-tunes a Llama3.1-8B-Instruct on \ours takes approximately 8 hours. We use PyTorch FSDP \citep{paszke2019pytorch} for efficient training.
\subsection{Perplexity Selection}
The perplexity-based selection method identifies critical steps by measuring the generation difficulty of each step within a trajectory. We compute perplexity as the exponentiated average negative log-likelihood of a step's tokens under the model's distribution. Formally, the perplexity for step \( x \) is defined as:

\begin{equation}
\text{PPL}(x) = \exp\left(-\frac{1}{|x|} \sum_{t=1}^{|x|} \log p(x_t \mid x_{1:t-1})\right)
\end{equation}

where \( |x| \) is the token length of the step, and \( p(x_t \mid x_{1:t-1}) \) is the model's probability for token \( x_t \) given its context. Higher perplexity values indicate steps that are more challenging for the model to generate.

Instead of using a fixed threshold, we dynamically select the top 30\% of steps with the highest perplexity within each trajectory. 

\subsection{Value Estimation} \label{app: value estimation}
We follow the rollout setup of \citep{xiong2024watch}, which defines the rollout estimate reward function as \(r_s(s_t, a_t)\) as the estimated outcome reward from the exploration starting step \(t\). The agent with policy \(\pi_s\) is employed to rollout trajectory \(\tau_{t:n}\) from step \(t\), based on the historical trajectory \(\tau_{t-1}\). The environment then gives a final reward \(r_o(i, \tau_n, E)\) for the trajectory. The estimated reward can be calculated as:
\begin{equation}
    r_s(s_t, a_t) = \mathbb{E}_{\tau_n \sim \pi_s(\tau_{t:n}|\tau_{t-1})}[r_o(i, \tau_n,E)]
\end{equation}
Due to the complexity of computing this expectation value directly, we utilize the Monte Carlo sampling technique for estimation. By drawing \(N\) trajectories from step \(t\) using \(\pi_s\), we create a collection of trajectories:
\begin{equation}
    \{\tau^{(i)} | i = 1, \ldots, N\} = \text{MC}_{\pi_s}(\tau_{t-1}; N)
\end{equation}
The estimated reward is then calculated as:
\begin{equation}
    r_s(s_t, a_t) = 
\begin{cases} 
\frac{1}{N} \sum_{i=1}^N r_o(i, \tau_n, E), & \text{for } t < n, \\
r_o(i, \tau_n, E), & \text{for } t = n.
\end{cases}
\end{equation}
Then the value function \( V^\pi(s_t) \) is estimated as the expectation over all possible trajectories starting from state \( s_t \) under policy \( \pi \):
\begin{equation}
    V^\pi(s_t) = \mathbb{E}_{\tau_{t:n} \sim \pi(\tau_{t:n} | s_t)} \left[ \sum_{k=t}^n \gamma^{k-t} r_s(s_k, a_k) \right],
\end{equation}
where \( \gamma \) is the discount factor which we set to be 0.99 here. 
Finally, we compute the difference between consecutive steps. If this difference exceeds the threshold of 0.1, the step is identified as a critical step. At this point, we designate the expert action corresponding to the same timestep $t$ as trainable and calculate the loss associated with that action.

\section{Additional Results} \label{app: add results}
Detailed results of other backbone models, random selection and different critical selection ratio $m$ are shown in Table \ref{tab: addition results}.
\begin{table*}[t]
\centering
\resizebox{0.9\textwidth}{!}{ 
\begin{tabular}{c|ccccccccccc|ccccc}
\hline
\multirow{2}{*}{\textbf{Model}} & \multicolumn{11}{|c|}{\textbf{held-in}} & \multicolumn{5}{|c}{\textbf{held-out}} \\ 
\cline{2-17}
& Alfworld & Babyai & Maze & Movie & Sciworld & Textcraft & Todo & Weather & Webshop & Wordle & AVG & Sheet & Academic &Jericho &PDDL & AVG \\ 
\hline
\multicolumn{14}{l}{\textit{Mistral-7B-Instruct-v0.3}} \\
\hline
AgentTraj-L & 79.50 & 76.43 & \textbf{44.00} & 75.00 & 47.32 & \textbf{76.00}  & 75.00 & 25.00 & \textbf{72.00} & \textbf{8.00} & 57.83 & 35.00 & 22.39 & \textbf{6.55} & 2.50 & 16.61\\
Ours & \textbf{85.00} & \textbf{77.73} & 40.00 & \textbf{85.00} & \textbf{56.59} & 69.00 & \textbf{85.00} & \textbf{30.00} & \textbf{72.00} & \textbf{8.00} & \textbf{60.83} & \textbf{45.00} & \textbf{30.44} &  6.32 & \textbf{2.53} & \textbf{21.07} \\
\hline
\multicolumn{14}{l}{\textit{Qwen2.5-7B-Instruct}} \\
\hline
AgentTraj-L &\textbf{78.50} & 73.48 &32.00 & \textbf{80.00} & 39.51 & 55.00 & \textbf{90.00} & \textbf{40.00} & \textbf{65.50} & 8.00 & 56.20 & \textbf{55.00} & \textbf{41.56} & 14.52 & 9.77 & 30.21 \\
Ours & 76.50&	\textbf{80.90}&	\textbf{40.00}&	\textbf{80.00}&	\textbf{41.32} & \textbf{64.00}&	80.00&	\textbf{40.00}&	62.50&	\textbf{12.00}&	\textbf{57.72} & \textbf{55.00} & 39.96 &\textbf{16.76} & \textbf{11.72} & \textbf{30.86}\\
\hline
\multicolumn{14}{l}{\textit{Random Selection}} \\
\hline
40\% & \textbf{83.50} & \textbf{73.08} & \textbf{60.00} & \textbf{85.00} & 37.51 & \textbf{69.00}  & 75.00 & 50.00 & \textbf{69.50} & 16.00 & \textbf{61.86} & 55.00 & 40.71 & \textbf{18.97} & 15.96 & 32.66\\
30\% & \textbf{83.50} & 71.70 & 40.00 & 80.00 & 38.32 & 67.00 & \textbf{80.00} & 50.00 & 68.50 & \textbf{20.00} & 59.90 & \textbf{70.00} & \textbf{48.63}& 14.80 & \textbf{18.71} & \textbf{38.04} \\
20\% & 81.00 & 72.30 & 36.00 & 80.00 & 39.98 & 65.00 & \textbf{80.00} &\textbf{60.00} & 64.50 & 12.00 & 59.08 & 65.00 & 43.35 & 13.93 & 16.64 & 34.73 \\
10\% & 78.50 & 72.78 & 32.00 & 80.00 & \textbf{41.02} & 66.00 & \textbf{80.00} & 55.00 & 65.50 & 16.00 & 58.68 & 55.00 & 36.95 & 9.80 & 14.39 & 29.04 \\
\hline
\multicolumn{14}{l}{\textit{Critical Selection Ratio}} \\
\hline
30\% &\textbf{84.50}& 80.98&	\textbf{48.00}&	\textbf{90.00}&	\textbf{42.01}&	\textbf{72.00}&	\textbf{90.00}&	\textbf{60.00}&	\textbf{71.50}&	\textbf{20.00}&	\textbf{65.91} & \textbf{70.00} & \textbf{49.39} & 18.21 & 15.84 & \textbf{38.36}\\
20\% & 82.50 & 73.77 & \textbf{48.00} & 80.00 & 34.68 & \textbf{72.00} & 80.00 & 50.00 & 71.00 & 12.00 & 60.40 & \textbf{70.00} & 43.90 & \textbf{25.01}& 8.94 & \textbf{36.96}\\
10\% & 78.00 & \textbf{83.76} & 44.00 & 75.00 & 36.13 & 64.00 & 80.00 & 50.00 & 67.00 & 12.00 & 58.99 & 60.00 & 35.63 & 13.01 & \textbf{16.08} & 31.18\\
\hline
\end{tabular}
}
\caption{Detailed performance of other backbone models, random selection and different critical selection ratio $m$. The best performance in each section is highlighted in \textbf{bold}.}
\label{tab: addition results}
\end{table*}
\newpage
\section{Prompts Details} 
\subsection{Prompts of Critical Step Selection} \label{a_c_prompt}
\begin{tcolorbox}[breakable, title = {Critcal Selection Prompts},boxsep=1pt, left=2pt, right=2pt, top=2pt, bottom=2pt, fonttitle=\small, fontupper=\small]
A critical step is defined as a key action or decision that, if performed correctly, significantly increases the likelihood of successfully completing the task. It represents a turning point in the process that influences the outcome of subsequent actions. More specifically, critical steps include:
\begin{itemize}
    \item \textbf{Plan Creation}: Steps where the LLM agent formulates sub-goals by analyzing previous observations and considering the final objective, breaking down the larger goal into manageable tasks that guide the agent's actions towards the overall outcome.
    \item \textbf{Critical Observation}: Steps where the LLM agent identifies and analyzes key information from the environment, which help agent understand the objective or state and refine its strategy and decision-making towards more effective outcomes.
    \item \textbf{Critical Action}: Steps where the LLM agent takes decisive and impactful actions based on prior observations, significantly advancing the process toward the final objectives. These actions are crucial in shaping the direction of the agent's strategy and are often pivotal moments that determine progress or failure, ensuring that the agent remains on track to achieve the desired outcome.
    \item \textbf{Self Correction}: Steps where the LLM agent carefully recalls and assesses its previous actions or decisions, especially after encountering failure or suboptimal outcomes. During this process, the agent reflects on what went wrong, identifies areas for improvement, and adjusts its approach to enhance future performance, which helps the agent refine its decision-making and better align with the overall objective.
\end{itemize}
\textbf{Task Description:} 
Your task is:
\begin{enumerate}
    \item Induce a high-level plan or strategy based on the expert conversation, summarizing the key steps needed to successfully complete the task.
    \item Based on this high-level plan, identify the most critical action steps in the expert conversation. A maximum of \{\} steps may be chosen from the conversation.
    \item Provide a detailed explanation for choosing these critical steps, specifying which category (e.g., key observation, planning, recall, pivotal action) they belong to and why mastering these steps ensures the success of the task.
\end{enumerate}
\textbf{Answer Format:}
\begin{enumerate}
    \item \textbf{The high-level plan is:} [Summarize the strategy and key steps for task completion]
    \item \textbf{The critical steps are:} conversation[...]
    \item \textbf{Reason:} [Explain why these steps are critical, including which category they fall into (key observation, planning, recall, pivotal action) and how they enable the player to avoid mistakes in subsequent steps]
\end{enumerate}
\end{tcolorbox}
\subsection{Prompts for Task Inference}
We adopt the setup of AgentGym \citep{xi2024agentgym} and utilize the same prompts for task inference. For further details, please refer to Appendix G: Prompt Details of AgentGym.

\end{document}